\documentclass[a4paper,fleqn,numbers,sort]{cas-dc}

\usepackage[numbers,sort&compress]{natbib}
\usepackage{algorithm}
\usepackage{algpseudocode}

\usepackage{booktabs}
\usepackage{tabularx}
\usepackage{multirow}
\usepackage{url}
\usepackage{amsmath}
\usepackage{graphicx}
\graphicspath{{figures/}}
\usepackage{subcaption}
\usepackage{placeins}
\usepackage{needspace}
\usepackage{xcolor}
\def\tsc#1{\csdef{#1}{\textsc{\lowercase{#1}}\xspace}}
\tsc{WGM}
\tsc{QE}

\begin{document}
\let\WriteBookmarks\relax
\hypersetup{hidelinks}
\makeatletter \setlength{\@fptop}{0pt} \setlength{\@fpsep}{12pt} \setlength{\@fpbot}{0pt plus 1fil} \makeatother \renewcommand{\topfraction}{0.95} \renewcommand{\bottomfraction}{0.95} \renewcommand{\textfraction}{0.06} \renewcommand{\floatpagefraction}{0.80} \renewcommand{\dbltopfraction}{0.95} \renewcommand{\dblfloatpagefraction}{0.80} \setlength{\textfloatsep}{12pt plus 3pt minus 3pt}\setlength{\dbltextfloatsep}{12pt plus 3pt minus 3pt} \setlength{\floatsep}{12pt plus 2pt minus 2pt} \setcounter{topnumber}{4} \setcounter{bottomnumber}{3} \setcounter{totalnumber}{6} \setcounter{dbltopnumber}{4}

\shorttitle{Delayed KV Eviction via Layer-Wise Chunked Prefill}

\shortauthors{Wang et al.}

\title [mode = title]{TideKV: Delayed KV Eviction via Layer-Wise Chunked Prefill for Memory-Efficient Long-Context LLM Inference}


%

\author[1]{Yu Wang}[orcid=0009-0006-4754-164X]
\ead{wangyu23@mail.nwpu.edu.cn}
\credit{Writing - review \& editing, Writing - original draft, Validation, Software, Methodology, Investigation, Formal analysis.}

\author[1]{Zhu Wang}
\cormark[1]
\ead{wangzhu@nwpu.edu.cn}
\credit{Writing-review \& editing, Writing – original draft, Methodology, Formal analysis, Data curation.}

\author[2]{Ning Hu}
\ead{hnhawk@163.com}
\credit{Supervision, Project administration, Funding acquisition, Conceptualization.}

\author[1]{Yifan Guo}
\ead{guoyf@mail.nwpu.edu.cn}
\credit{Methodology, Investigation.}

\author[2]{Xiu Ma}
\ead{2750162952@qq.com}
\credit{Methodology, Investigation.}

\author[1]{Bin Guo}
\ead{guob@nwpu.edu.cn}
\credit{Supervision, Project administration, Funding acquisition, Conceptualization.}

\author[1,3]{Zhiwen Yu}
\ead{zhiwenyu@nwpu.edu.cn}
\credit{Supervision, Project administration, Funding acquisition, Conceptualization.}





\affiliation[1]{organization={Northwestern Polytechnical University},
addressline={No. 127 West Youyi Road}, city={Xian},
postcode={710129}, state={Shaanxi}, country={China}}

\affiliation[2]{organization={Xi'an Aeronautics Computing Technique              Research Institute, AVIC},
addressline={No. 15 Jinye 2nd Road}, city={Xi'an}, postcode={710065}, state={Shaanxi}, country={China}}

\affiliation[3]{
organization={College of Computer Science and Technology, Harbin Engineering University}, addressline={No. 145 Nantong Street}, city={Harbin}, postcode={150001}, state={Heilongjiang}, country={China} }







\cortext[1]{Corresponding author}



\begin{abstract}[]
Long-context large language model (LLM) inference is important for modern intelligent applications but remains constrained by peak memory during prompt prefill. Existing KV cache compression reduces decoding memory but typically requires full-prompt prefill to preserve global context, retaining a high construction peak. Online chunk-level eviction lowers memory earlier but selects before each layer has observed the complete prompt, potentially degrading cache construction and decoding. We present TideKV, an on-GPU cache-construction architecture using layer-major delayed eviction. TideKV processes all prompt chunks within each Transformer layer, retains full-prompt KV only for the active layer, and evicts at layer end. This schedule preserves full-prompt-aware selection while reducing activation and temporary memory. {\color{black} Experiments across multiple computing platforms show that TideKV preserves the inference quality and final cache size of the evaluated compression methods.} At 64K on A100, it reduces the construction-time transient footprint by 83.89\%. {\color{black} On RTX 4090, it increases time to first token by only 1.06\% at 64K and enables 128K prefill.} TideKV also completes 128K on Jetson, whereas matched post-prefill compression runs out of memory on both resource-constrained platforms. These results show that delayed layer-end eviction substantially expands feasible context lengths without sacrificing cache utility.
\end{abstract}






\maketitle


\section{Introduction}
\label{sec:intro}
Recent advances in large language models (LLMs) have extended supported context windows to 128K tokens and beyond~\cite{grattafiori2024llama,team2024gemini,hui2024qwen2}, enabled in part by positional-encoding extensions such as rotary embeddings~\cite{su2024roformer,peng2024yarn,ding2024longrope}. This expansion is also driven by applications that require sustained reasoning over long horizons, including retrieval-augmented generation~\cite{lewis2020retrieval}, multi-turn autonomous agents~\cite{yao2023react}, long-form code synthesis~\cite{roziere2023code}, and extended document comprehension. Such applications routinely require tens of thousands of input tokens in a single inference session and have motivated dedicated long-context benchmarks~\cite{bai2024longbench,hsieh2024ruler}. At the same time, due to requirements related to data privacy, communication latency, and regulatory compliance, the deployment of LLMs is undergoing a structural shift from centralized cloud infrastructure to edge computing platforms~\cite{xu2024device,alizadeh2024llm,song2024powerinfer}, such as mobile robots, embodied intelligent systems~\cite{driess2023palm,zitkovich2023rt}, and in-vehicle autonomous driving modules~\cite{zhu2024survey}.

However, resource-constrained edge accelerators operate under fixed and often severely limited device memory, creating a fundamental tension with the memory demands of long-context inference. During inference, memory is consumed by multiple concurrent components: the key--value (KV) cache accumulating per-layer attention states, the full-sequence hidden-state tensor propagated through Transformer blocks~\cite{vaswani2017attention}, intermediate MLP activation tensors, and per-layer attention score matrices. As input sequence length increases, each of these components grows correspondingly, with the KV cache scaling linearly with both sequence length and layer count~\cite{ainslie2023gqa}. The aggregate memory pressure renders long-context inference on resource-constrained edge devices increasingly infeasible without systematic memory management. The memory pressure is most acute during the prefill phase, when the entire input prompt must be processed to construct the initial KV cache before decoding begins~\cite{agrawal2024taming}. For a representative Transformer model such as Llama-3.1-8B~\cite{grattafiori2024llama}, {\color{black} a single 128K-token prompt generates approximately 16~GiB of uncompressed KV cache.} {\color{black} During prefill, hidden states and temporary activations coexist in device memory with the KV cache}~\cite{dao2024flashattention}. The cumulative footprint routinely exceeds the physical memory capacity of edge accelerators, resulting in out-of-memory (OOM) errors that render the service unavailable.

\begin{figure*}[pos=t]
\centering
\includegraphics[width=\textwidth]{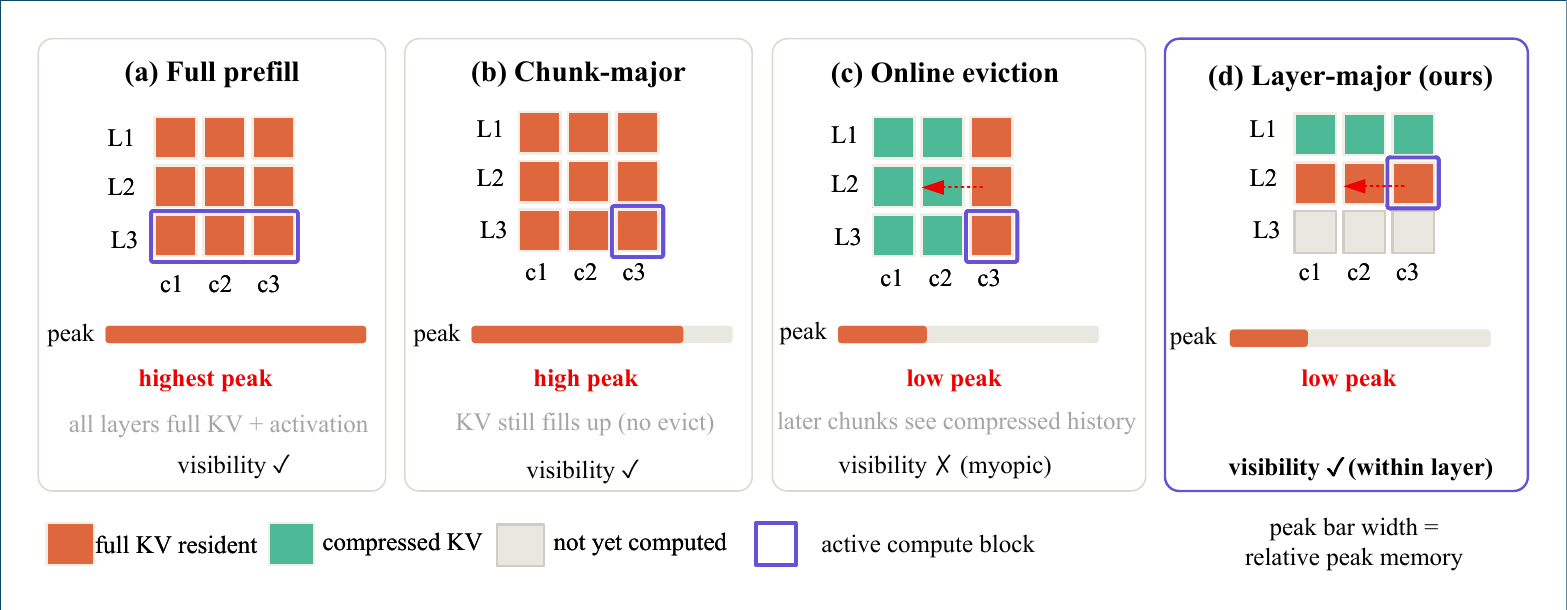}
\caption{Four schedules for constructing a compressed KV cache during prefill. (a) {\color{black} Full prefill without intermediate cache release retains uncompressed KV across all layers until the prompt has been processed, preserving full-prompt visibility at a high construction peak.} (b) Chunked prefill without eviction processes the prompt chunk by chunk, yet every chunk's KV stays resident, so memory still grows with prompt length. (c) Chunked prefill with online eviction evicts after each chunk to cap memory, but it decides what to keep before the full prompt is seen, {\color{black} which can degrade quality through myopic eviction.} (d) TideKV (ours) streams the prompt chunk by chunk in a layer-major order and defers eviction to each layer's boundary, so every layer scores against the full prompt {\color{black} while full-prompt KV is confined to one active layer and temporary activations to one chunk, alongside the hidden-state bridge and retained caches.}}
\label{fig:concept}
\end{figure*}

Therefore, to enable long-context reasoning on resource-constrained devices, it is necessary not only to control the final KV cache retained during decoding but also to limit the peak memory consumption during the construction of the KV cache. Existing KV cache compression approaches fall into two paradigms, each exposing a distinct limitation under tight memory budgets, as illustrated in Figure~\ref{fig:concept}. {\color{black} Post-prefill selectors such as SnapKV~\cite{li2024snapkv} score and compress each layer's cache after processing the complete prompt at that layer.} While this preserves full visibility of the prompt during the selection process, it still results in {\color{black} high construction-memory peaks} {\color{black} because each layer still processes the whole prompt at once before compressing its cache}. In contrast, {\color{black} online methods such as KeyDiff~\cite{park2026keydiff} evict KV during prompt processing to control memory usage.} {\color{black} Because these methods evict tokens before observing the complete prompt, they may prematurely remove tokens whose importance depends on later context. This myopic eviction can reduce compression quality.} This creates a central visibility-residency trade-off: current methods either spend substantial memory to preserve full-prompt information, or bound memory by evicting before the prompt is fully observed, which can compromise cache quality.

{\color{black} Full-prompt visibility can be preserved while lowering construction-time peak memory by reordering prefill execution.} When a Transformer layer is driven to completion over the whole prompt before the next layer begins, and eviction is deferred to that layer's boundary, every layer scores its tokens against the uncompressed causal history of the entire prompt, {\color{black} so selection is not preceded by eviction of the layer's causal prompt history.} At the same time, full-prompt KV is resident for only one layer at a time. {\color{black} Streaming the prompt in chunks limits temporary attention/MLP activations to one chunk,} which bounds the activation workspace that dominates the prefill peak. {\color{black} Relative to an uncompressed cache spanning all $N$ layers, this schedule reduces uncompressed KV residency from $O(NS)$ to $O(S)$ while preserving full-prompt visibility.}

We present TideKV, a prefill execution architecture that realizes this principle through layer-major delayed eviction. For each layer, TideKV streams the prompt chunk by chunk against that layer's accumulated, uncompressed KV history; only after the layer has consumed the whole prompt does it compress the layer's KV to the target budget and release the full KV before advancing. The compression step uses a two-pass blockwise streaming top-$k$ procedure: a first pass accumulates log-sum-exp normalization statistics over key blocks, and a second pass derives the retained-token selection, so that scoring never materializes the full observation-by-prefix attention matrix. This scoring step is a replaceable component: because the schedule is decoupled from the token-importance rule, existing KV-cache compression selectors can be used in its place. TideKV further supports layer- and head-adaptive budget allocation within the same delayed-eviction invariant, enabling fine-grained memory--quality control across heterogeneous attention configurations. All state remains on the GPU, and the output is a compressed initial decode cache.

The main contributions of this work are as follows:
\begin{itemize}
{\color{black} \item We propose \emph{layer-major delayed eviction}, which preserves full-prompt visibility for KV selection while confining uncompressed KV to one active layer and temporary attention/MLP activations to one chunk. Eviction occurs only after the layer finishes prompt encoding.}
\item {\color{black} We separate construction scheduling from \emph{prefix-based} selectors compatible with layer-end selection and delayed eviction, and evaluate four selector families within this framework.} To keep the compression step itself within budget, we design a memory-controlled two-pass blockwise streaming top-$k$ procedure whose temporary footprint is independent of prompt length.
\item We evaluate TideKV on Llama and Mistral models over the LongBench and RULER benchmarks together with the memory and feasibility of long-context prefill. At matched budgets it preserves downstream quality and final-cache size. At 64K on A100, TideKV reduces the construction-time transient footprint by 83.89\%; {\color{black} on the RTX 4090, the time-to-first-token increase is 1.06\% at 64K.} It completes 128K-token prefill on both resource-constrained platforms where matched post-prefill compression is out of memory. \end{itemize}

The remainder of this paper is organized as follows. Section~\ref{sec:related} reviews related work. Section~\ref{sec:problem} defines the problem setting. Sections~\ref{sec:framework} and~\ref{sec:impl} present the TideKV framework and implementation. Section~\ref{sec:eval} reports the evaluation, and Section~\ref{sec:conclusion} concludes the paper.

\section{Related work}
\label{sec:related}

In this section, we review the relevant research in the following three aspects:

\subsection{{\color{black} KV-cache selection and compression}}
\label{subsec:postprefill}
{\color{black} KV-cache compression methods differ in which entries they retain and when they make eviction decisions.} Heavy-hitter eviction keeps tokens with high accumulated attention together with recent tokens~\cite{zhang2023h2o,liu2023scissorhands}; StreamingLLM retains a few initial attention-sink tokens plus a sliding window so that a bounded cache supports streaming beyond the nominal context~\cite{xiao2024efficient}; SnapKV predicts from a prompt-final observation window which positions generation will attend to and compresses per head~\cite{li2024snapkv}; and FastGen profiles the per-head attention structure to choose a per-head compression policy~\cite{ge2024model}. A parallel line refines the granularity of the budget: PyramidKV allocates a pyramidal per-layer budget~\cite{cai2024pyramidkv}; Ada-KV distributes budget across heads~\cite{feng2026ada}; CAKE casts eviction as adaptive layer-preference budgeting with time-varying token importance~\cite{qin2025cake}; ChunkKV retains semantic chunks rather than isolated tokens~\cite{liu2026chunkkv}. In contrast, query-aware methods load only the KV pages required by the query during the decoding phase~\cite{tang2024quest}. Beyond hard eviction, several methods modify or complement it: KeepKV merges KV entries to limit output perturbation~\cite{tian2026keepkv}; KVzip targets query-agnostic caches reusable across queries~\cite{kim2026kvzip}; a recent direction reconstructs KV on demand from a stored residual stream~\cite{qasim2026residual}; and, orthogonally to eviction, low-bit quantization compresses each retained entry rather than removing it~\cite{liu2024kivi}. Surveys catalog these directions~\cite{shi2024keep}.

{\color{black} Post-prefill selectors preserve full-prompt visibility with a compact retained cache; layer- and head-adaptive allocation also informs TideKV's adaptive variants. Their bottleneck is cache construction. Even if layer-wise eviction confines uncompressed KV to one layer, whole-prompt execution still produces full-sequence hidden states and MLP activations, driving the prefill peak (Section~\ref{subsec:prefillpeak}). Thus, constructing a compact cache can exceed device memory (Fig.~\ref{fig:concept}a). TideKV retains the selector and changes compression scheduling to reduce this construction peak.}

\subsection{Chunked construction with online eviction}
\label{subsec:chunked}
{\color{black} Online construction processes prompts in chunks and evicts before full-prompt observation (Fig.~\ref{fig:concept}c). InfiniPot distills history as the cache fills, processing later input against compressed history under a fixed budget~\cite{kim2024infinipot}; KeyDiff uses key similarity for FlashAttention-compatible eviction under a fixed cache budget~\cite{park2026keydiff}; EpiCache clusters conversational history into episodes with layer-budgeted eviction and highlights the peak cost of full-context prefill~\cite{kim2024epicache}. Related designs use look-once multimodal eviction~\cite{wan2024look} and trained retaining heads~\cite{huang2024locret}. Early eviction lowers construction peaks, but later tokens see compressed history and may lose entries relevant to subsequent context. TideKV delays eviction until each layer finishes, preserving uncompressed causal history during prompt encoding and full-prompt visibility for selection.}

\subsection{Prefill peak-memory reduction and KV offloading}
\label{subsec:prefillpeak}
{\color{black} A third group of methods targets peak prefill memory by reducing intermediate MLP activations, which can dominate the memory footprint}~\cite{zhang2025mom} (Fig.~\ref{fig:concept}b). Chunked prefill splits the prompt into segments processed sequentially. {\color{black} For prompt length $S$ and chunk size $C$, chunking reduces the corresponding temporary activation term to approximately $C/S$ of its full-sequence size, while introducing kernel-launch and latency overheads}~\cite{agrawal2023sarathi,yu2022orca,zhong2024distserve}; MST partitions the MLP and LM-head inputs into mini-sequences during training~\cite{luo2024mini}, and MOM brings mini-sequence MLP processing to inference while offloading the full KV cache to CPU~\cite{zhang2025mom}; {\color{black} PyramidInfer instead computes fewer KV entries in deeper layers during prefill}~\cite{yang2024pyramidinfer}. Offloading-centric systems keep the full KV but relocate it: HeadInfer fetches KV head-group by head-group between CPU and GPU so that an entire layer's KV never co-resides on the GPU, at a substantial decode-speed cost~\cite{luo2025headinfer}; FlexGen aggregates GPU, CPU, and disk memory to run inference under tight GPU budgets~\cite{sheng2023flexgen}. Collectively, these methods reduce the peak load during the prefill phase but either retain the uncompressed decoding cache or incur the cost of data transfer between the GPU and CPU. TideKV reduces peak memory usage through chunk-inner execution {\color{black} while keeping execution on the GPU} and generating a compressed decoding cache.

{\color{black} The approaches differ in prompt visibility and memory use during construction. Post-prefill selectors preserve full-prompt visibility but mainly reduce retained KV; online eviction also lowers the construction peak while exposing later chunks to compressed history. Activation partitioning and offloading reduce prefill residency but may retain the uncompressed decode cache or move KV to the CPU. TideKV combines layer-major execution with delayed layer-end eviction: full-prompt KV is confined to one active layer, temporary activations to one chunk, and selection observes the complete prompt. Construction stays on the GPU.}

\section{Problem formulation}
\label{sec:problem}

This section formalizes the compressed-cache construction problem that TideKV addresses. Our concern is not the design of a particular token-importance rule, but the \emph{execution setting} in which a compressed initial decode cache is constructed from a long prompt under a device memory budget. The formulation deliberately separates two questions that prior work often couples: \emph{which} KV entries to retain (the \emph{selection} problem), and in \emph{what order} prefill is executed and \emph{when} eviction takes effect (the \emph{scheduling} problem). {\color{black} TideKV specifies the construction schedule and leaves the selector configurable.}

\subsection{Notation and setting}

{\color{black} We consider a pretrained decoder-only Transformer (Table~\ref{tab:notation}) processing one prompt $x_{1:S}$ in chunks of size $C$. Prefill constructs each layer's KV; decoding uses the retained cache. We fix the model weights and treat selection as replaceable to isolate cache-construction cost.}

\begin{table}[pos=t]
\centering
\caption{Notation.}
\label{tab:notation}
\small
\begin{tabularx}{\columnwidth}{@{}lX@{}}
\toprule
Symbol & Meaning \\
\midrule
$N$ & number of Transformer layers \\
$B$ & batch size \\
$S$ & prompt (sequence) length \\
$C$ & chunk size used during prefill \\
$H_q, H_{kv}$ & number of query heads, KV heads \\
$D$ & head dimension \\
$b$ & element size in bytes \\
$W$ & observation-window size used for scoring \\
$K_l$ & retained cache budget for layer $l$ \\
$d_{\text{model}}$ & model hidden dimension ($=H_q\cdot D$) \\
$\hat{\mathcal{C}}$ & compressed initial decode cache (the output) \\
$\Pi$ & {\color{black} construction procedure with a specified schedule, selector, and budget} \\
$d_{\text{ff}}$ & MLP intermediate width ($=O(d_{\text{model}})$) \\
\bottomrule
\end{tabularx}
\end{table}

\subsection{Memory quantities in cache construction}

A prompt KV cache occupies $2NBH_{\mathrm{kv}}SDb$ bytes. This expression counts stored KV alone; it does not describe either the final compressed cache or the peak memory required to construct it. We therefore distinguish three memory quantities. \begin{enumerate}
\item \textbf{Final compressed cache size}: $S_{\text{final}}(\Pi)$, the number of retained KV positions after compression; under a per-layer budget $K_l$ it equals the total nominal budget $K_{\text{tot}}\triangleq\sum_l K_l$, while its byte footprint follows after adjustment for the number of KV heads and the cache layout. \item \textbf{Construction-time live-tensor peak}: the part of the prefill footprint directly controlled by the cache construction procedure, \begin{equation}
\label{eq:peak-memory}
\begin{aligned}
M_{\mathrm{live}}(\Pi)=\max_l\bigl[&
\mathrm{KV}_{\mathrm{res}}(l)+\mathrm{Hidden}(l)\\
&+\mathrm{Act}(l)+\mathrm{Temp}(l)\bigr].
\end{aligned}
\end{equation}
Here, $\mathrm{KV}_{\mathrm{res}}$ is resident prompt KV, $\mathrm{Hidden}\propto B S d_{\mathrm{model}}$ is the hidden-state bridge, $\mathrm{Act}$ is the active attention/MLP workspace, and $\mathrm{Temp}$ contains transient scoring, selection, and cache-packing tensors. The observed process peak additionally contains terms not determined solely by $\Pi$:
\begin{equation}
\label{eq:process-memory}
\begin{aligned}
{\color{black} M_{\mathrm{proc}}\approx{}}&M_{\mathrm{model}}+M_{\mathrm{live}}(\Pi)
+M_{\mathrm{backend}}\\
&+M_{\mathrm{allocator}}+M_{\mathrm{external}}.
\end{aligned}
\end{equation}
where $M_{\mathrm{backend}}$ includes kernel workspaces, masks, and attention buffers; $M_{\mathrm{allocator}}$ includes reserved-but-inactive blocks and fragmentation; and $M_{\mathrm{external}}$ covers CUDA-context, page-table, runtime-metadata, and direct allocations not attributed to active PyTorch tensors. Together, Eqs.~\eqref{eq:peak-memory} and~\eqref{eq:process-memory} separate the schedule-dependent working set from runtime memory contributed by the model, kernels, and memory manager. {\color{black} Equation~\eqref{eq:process-memory} is a schematic accounting relation; component peaks need not coincide. This decomposition underlies the live-allocation and process-level measurements} in Section~\ref{sec:eval}. \item \textbf{Temporary compression memory} $\mathrm{Temp}(\Pi)$: score matrices, selection buffers, and attention intermediates. A naive layer-end compression that materializes a $W\times S$ observation-by-prefix score matrix can make $\mathrm{Temp}$ dominate $M_{\mathrm{live}}$.
\end{enumerate}

Bounding $\mathrm{KV_{res}}$ alone does not bound the construction peak because $\mathrm{Hidden}$ and $\mathrm{Act}$ also grow with $S$. Allocator retention and backend workspaces further determine how a smaller live working set appears in reserved and process-level memory. TideKV therefore bounds $M_{\mathrm{live}}$ and $\mathrm{Temp}$ while retaining the same $S_{\mathrm{final}}$ as its underlying selector. Section~\ref{sec:eval} reports allocated, reserved, and process-level memory under multiple allocators, attention backends, and persistent workloads. \begin{figure*}[pos=htbp]
    \centering
    \scalebox{1}[0.8]{%
        \includegraphics[width=\textwidth]{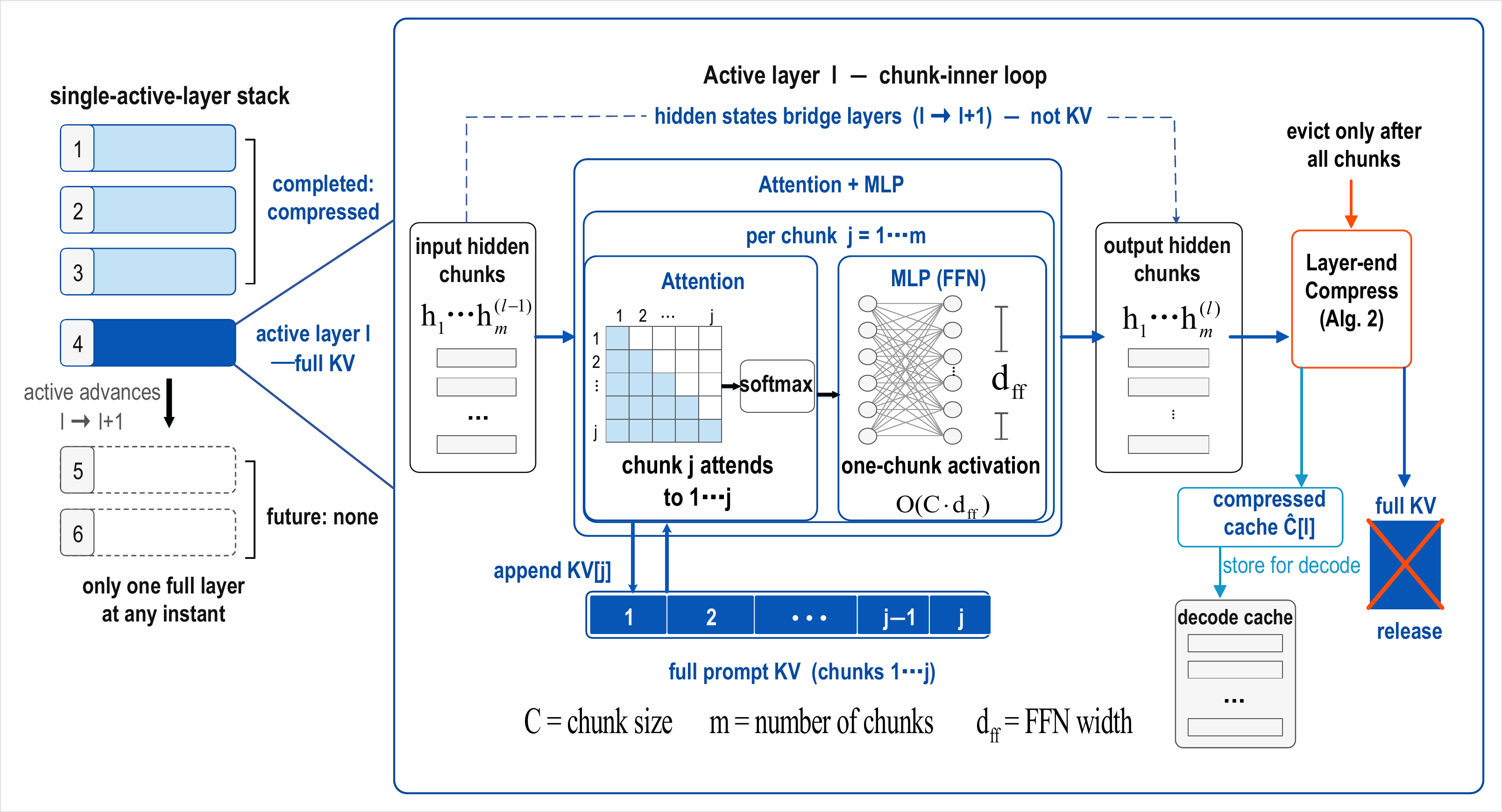}%
    }
\caption{Layer-major delayed eviction and the corresponding tensor lifetimes. \emph{Left:} full prompt KV is resident only for the active layer $l$; completed layers hold compressed caches and future layers hold none. \emph{Right:} chunk $j$ attends to the accumulated uncompressed KV of chunks $1,\ldots,j$, while output hidden chunks form the bridge to layer $l{+}1$. After all $m$ chunks complete, TideKV gathers the retained KV and absolute-position metadata into $\hat{\mathcal{C}}[l]$ and releases the full KV. Hidden-state write/read and retained-KV gather are the principal added data movements; the layer-end commit is the synchronization boundary before layer $l{+}1$.}
\label{fig:lmde-flow}
\end{figure*}

\subsection{Construction objective}
\label{subsec:construction-goal}

{\color{black} A construction procedure $\Pi$ takes a tokenized prompt $x_{1:S}$, the fixed model, a chunk size $C$, and a selector with its cache-budget specification. It returns a compressed initial decode cache $\hat{\mathcal C}$ and the prefill output needed to start generation. For a fixed selector $F$ and layer/head budget allocation, our objective is to lower construction memory while preserving the prompt information available to $F$.

The reference procedure applies $F$ after uncompressed causal prefill. Let $Q_{\mathrm{full}}$, $Q_F$, and $Q_{\mathrm{Tide}}$ denote downstream scores for full-cache decoding, the selector-matched reference, and TideKV, respectively. On the same evaluation,
\begin{equation}
\label{eq:quality-decomposition}
Q_{\mathrm{full}}-Q_{\mathrm{Tide}}
=(Q_{\mathrm{full}}-Q_F)+(Q_F-Q_{\mathrm{Tide}}).
\end{equation}
The first term reflects selector compression; the second measures the additional change associated with construction scheduling. Algorithm~\ref{alg:lmde} implements layer-major execution and delayed eviction with explicit residency properties. Section~\ref{sec:framework} analyzes its prompt dependencies and working set; Section~\ref{sec:eval} evaluates fidelity and selector-matched task quality.}

\section{The TideKV framework}
\label{sec:framework}

Compressed-cache construction requires an execution schedule that bounds the live working set $M_{\mathrm{live}}$ and compression workspace $\mathrm{Temp}$ without sacrificing the full-prompt visibility needed for accurate selection. TideKV enforces a \emph{layer-major residency discipline}: it streams the complete prompt through one layer before advancing and delays eviction until the layer boundary. This schedule is governed by the \emph{single-active-layer residency invariant}: \begin{quote} At every instant of prefill, full-prompt KV is resident for at most one layer (the active layer), {\color{black} temporary attention/MLP activations are limited to one chunk alongside a prompt-wide hidden-state bridge}, and every completed layer holds only its compressed cache.
\end{quote}

{\color{black} \noindent This invariant confines uncompressed KV to one active layer and temporary activations to one chunk, while the hidden-state bridge and completed layers' compressed caches remain on the GPU. Delaying eviction until the layer boundary preserves full-prompt visibility for selection.} The framework rests on three mechanisms, detailed below and illustrated in Figs.~\ref{fig:lmde-flow} and~\ref{fig:compress-flow}: layer-major chunked prefill that gives each layer full causal visibility (\ref{subsec:core}), delayed layer-end eviction that confines full KV to the active layer (\ref{subsec:core}), and a memory-controlled compression step that keeps the eviction temporaries bounded (\ref{subsec:compress}).

\subsection{Layer-major delayed eviction}
\label{subsec:core}


As shown in Fig.~\ref{fig:lmde-flow}, TideKV recasts prefill as a \emph{layer-sequential} execution schedule: a standard forward pass propagates each chunk through all $N$ layers, whereas TideKV advances a single layer over all chunks, moving to the next layer only after the current one has consumed the entire prompt. During prefill it maintains three classes of state: completed layers hold \emph{compressed} caches, the \emph{active} layer holds full prompt KV for the chunks processed so far (the only full prompt KV resident anywhere), and future layers hold \emph{none}. Hidden states bridge the layers: after the active layer processes chunk $j$, {\color{black} its output hidden chunk $h_j^l$ replaces the consumed input chunk in the bridge; the resulting $H_l$ becomes the chunked input to layer $l{+}1$}. Because only the hidden outputs, not the KV, are needed to continue the forward computation, the active layer's full KV can be compressed and \textbf{released at the layer boundary}.

For layer $l$ with input chunks $H_{l-1}=\{h_1^{l-1},\dots,h_m^{l-1}\}$, $m=\lceil S/C\rceil$, TideKV processes chunks in causal order; when chunk $j$ is processed, its queries attend to the KV produced by chunks $1,\dots,j$ of the \emph{same} layer, so later chunks see the same uncompressed layer-local history a full-prompt execution would provide. Only after all $m$ chunks are processed does TideKV compress the layer's KV to its budget $K_l$, store the compressed cache, and release the full KV. Eviction therefore never alters the hidden states the layer has already produced, decoupling \emph{what is retained for decoding} from \emph{what is visible during prompt encoding}.

\begin{algorithm}[t]
\caption{Layer-major chunked prefill with delayed eviction}
\label{alg:lmde}
\begin{algorithmic}[1]
\Require {\color{black} hidden chunks $H_0=\{h_1^0,\dots,h_m^0\}$; layers $L_1,\dots,L_N$; budgets $K_1,\dots,K_N$; window $W$}
\Ensure {\color{black} compressed initial decode cache $\hat{\mathcal C}$; final hidden chunks $H$}
\State {\color{black} $\hat{\mathcal C}\gets\varnothing$; $H\gets H_0$}
\For{$l=1$ to $N$}
  \State {\color{black} $\mathrm{KV_{full}}\gets\varnothing$; $Q_{\mathrm{obs}}\gets\varnothing$}
  \For{$j=1$ to $m$} \Comment{causal chunk order}
    \State {\color{black} $H[j],\mathrm{KV_{full}}\gets L_l.\textsc{forward}(H[j],\mathrm{KV_{full}})$}
    \Comment{{\color{black} replace the consumed hidden chunk}}
    \State {\color{black} retain the latest $W$ queries in $Q_{\mathrm{obs}}$}
  \EndFor
  \State {\color{black} $\hat{\mathcal C}[l]\gets\textsc{LayerEndCompress}(\mathrm{KV_{full}},Q_{\mathrm{obs}},K_l)$}
  \State release $\mathrm{KV_{full}}$ \Comment{delayed eviction}
\EndFor
\State {\color{black} \Return $\hat{\mathcal C},H$}
\end{algorithmic}
\end{algorithm}

The pseudocode separates the \textbf{schedule} from both the \textbf{selector} inside \textsc{LayerEndCompress} and the \textbf{budget schedule} $\{K_l\}$. A SnapKV-style observation-window selector, a recent-token selector, or a head-adaptive selector {\color{black} can be used within the same scheduling framework}; the budget can be uniform ($K_l=K_{\text{tot}}/N$) or allocated adaptively across layers (more capacity to harder-to-compress layers) or heads. {\color{black} These choices preserve the uncompressed-KV residency invariant;} they change only which entries are retained and how the retained cache is laid out, with a head-adaptive budget yielding a variable-length cache stored in a flattened layout with per-head offsets.

{\color{black} The selector and budget schedule are configurable within this layer-end interface. Direct integration requires preserving full layer-local history until selection; compression algorithms with other execution or cache-representation requirements need additional adaptation.} Reordering execution confines full-prompt KV to one active layer and {\color{black} temporary attention/MLP activations to one chunk}, without changing the active layer's $O(S)$ KV term or its causal attention computation. Section~\ref{subsec:analysis} quantifies the resulting working set, and Section~\ref{sec:eval} evaluates the selector and allocation variants. \subsection{Memory-controlled layer-end compression}
\label{subsec:compress}

\begin{figure*}[pos=htbp]
    \centering
    \includegraphics[width=0.9\textwidth]{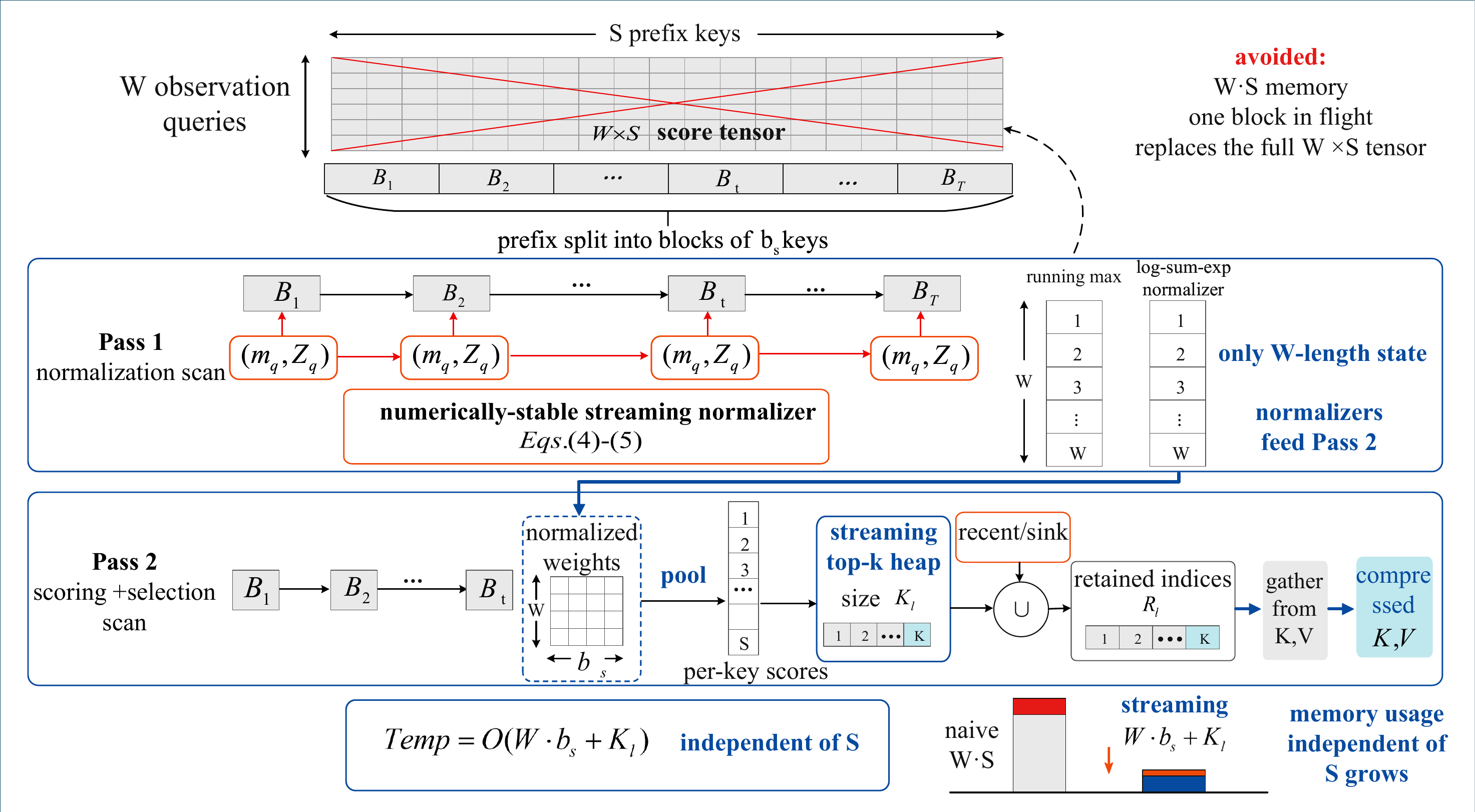}
    \caption{Memory-controlled two-pass blockwise scoring. Instead of materializing the dense $W\times S$ observation-by-prefix score tensor, the prefix keys are split into blocks. \emph{Pass 1} scans the blocks to build the numerically stable log-sum-exp normalizers $(m_q,Z_q)$, keeping only $W$-length state. \emph{Pass 2} rescans to form normalized weights, pools them into per-key scores, and {\color{black} keeps the top $K_l-|\mathcal R_{\mathrm{mand}}|$ nonmandatory entries together with the mandatory recent/sink set} before gathering the retained $(\hat K,\hat V)$. The temporary footprint $\mathrm{Temp}=O(W b_s+K_l)$ is independent of $S$.}
    \label{fig:compress-flow}
\end{figure*}

As shown in Fig.~\ref{fig:compress-flow}, the layer-end step must itself respect the invariant: a naive selector that scores $W$ observation queries against all $S$ prefix keys materializes a tensor of shape $[B,H_q,W,S]$, {\color{black} whose $O(B H_q W S)$ footprint can dominate $M_{\mathrm{live}}$.} TideKV avoids it with a \textbf{two-pass blockwise} procedure that streams over key blocks and never stores the full score tensor. Partition the prefix keys into blocks $\mathcal{B}_1,\dots,\mathcal{B}_T$ of size up to $b_s$. For observation query $q$ and prefix key $i$, the attention logit is

\begin{equation}
\label{eq:logit}
\ell_{q,i} = \frac{Q_q \cdot K_i}{\sqrt{D}}.
\end{equation}

{\color{black} For each observation query, logits for keys beyond its causal position are set to $-\infty$. Both passes use the same absolute positions and causal mask.}

\textbf{Pass 1} (normalization) {\color{black} uses an online running-maximum recurrence to accumulate normalization statistics without retaining logits from previous blocks:}
\begin{align}
m_q^{(t)} &= \max\!\big(m_q^{(t-1)},\ \max_{i\in\mathcal{B}_t}\ell_{q,i}\big), \label{eq:p1max}\\*
Z_q^{(t)} &= Z_q^{(t-1)}e^{\,m_q^{(t-1)}-m_q^{(t)}} + \sum_{i\in\mathcal{B}_t} e^{\,\ell_{q,i}-m_q^{(t)}}. \label{eq:p1norm}
\end{align}

After $T$ blocks, $M_q=m_q^{(T)}$ and $Z_q=Z_q^{(T)}$; {\color{black} Pass 1 stores $O(W)$ scalars per batch and attention head.}

\textbf{Pass 2} (scoring + streaming top-$k$) revisits the blocks, forms normalized weights, pools them into per-key importance, {\color{black} and keeps the top $k=K_l-|\mathcal R_{\mathrm{mand}}|$ nonmandatory entries:}
\begin{equation}
\label{eq:score}
a_{q,i} = \frac{e^{\,\ell_{q,i}-M_q}}{Z_q},\qquad
p_i = \sum_{q=1}^{W} w_q\, a_{q,i},
\end{equation}
\begin{equation}
\label{eq:retain}
{\color{black} \begin{aligned}
\mathcal R_l &= \operatorname{Top}_k\!\big(\{p_i:i\notin\mathcal R_{\mathrm{mand}}\}\big)
\cup\mathcal R_{\mathrm{mand}},\\
|\mathcal R_l|&=K_l.
\end{aligned}}
\end{equation}
{\color{black} where $w_q$ are pooling weights and $\mathcal{R}_{\text{mand}}$ is the set of always-retained token indices, such as recent tokens and attention sinks,} counted \emph{within} the budget $K_l$ rather than added on top, so the physical cache size equals the stated budget. {\color{black} The retention count assumes a feasible budget $|\mathcal R_{\mathrm{mand}}|\le K_l\le S$.} The compressed cache $\hat{\mathcal{C}}[l]$ gathers the rows indexed by $\mathcal{R}_l$ (or per-head $\mathcal{R}_{l,h}$).

\begin{algorithm}[t]
\caption{\textsc{LayerEndCompress} (two-pass blockwise scoring)}
\label{alg:compress}
\begin{algorithmic}[1]
\Require full layer KV $(K,V)$; observation queries $Q$ (size $W$); budget $K_l$; block size $b_s$; {\color{black} mandatory indices $\mathcal R_{\mathrm{mand}}$}
\Ensure retained index set $\mathcal{R}_l$ and gathered $(\hat K,\hat V)$
\State {\color{black} initialize $m_q\gets-\infty$, $Z_q\gets0$ for all $q$}
\For{each key block $\mathcal{B}_t$} \Comment{Pass 1: streaming log-sum-exp, \eqref{eq:logit}--\eqref{eq:p1norm}}
  \State update running max $m_q$ and normalizer $Z_q$ for all $q$
\EndFor
\State {\color{black} $k\gets K_l-|\mathcal R_{\mathrm{mand}}|$; $\mathrm{heap}\gets\varnothing$}
\For{each key block $\mathcal{B}_t$} \Comment{Pass 2: streaming top-$k$, \eqref{eq:score}--\eqref{eq:retain}}
  \State compute $a_{q,i}$ with $(M_q,Z_q)$; pool to $p_i$ for $i\in\mathcal{B}_t$
  \State {\color{black} retain top-$k$ candidates with $i\notin\mathcal R_{\mathrm{mand}}$ in heap}
\EndFor
\State $\mathcal{R}_l \gets \mathrm{indices}(\mathrm{heap}) \cup \mathcal{R}_{\text{mand}}$
\State \Return $\mathcal{R}_l,\ \textsc{gather}(K,V,\mathcal{R}_l)$
\end{algorithmic}
\end{algorithm}

{\color{black} The procedure leaves causal attention unchanged and bounds the} \emph{temporary} compression memory to $\mathrm{Temp}=O\!\big(B H_q (W b_s + K_l)\big)$, independent of $S$, compared with $O(B H_q W S)$ for the dense tensor. The second blockwise pass recomputes the logits in Eq.~\eqref{eq:logit} instead of storing them, and this work is included in the prefill overhead reported in Section~\ref{sec:eval}. The running maximum $m_q$ keeps the streaming log-sum-exp numerically stable in BF16 by rescaling every block to a common reference. Because each pass holds only one key block of size $b_s$, the compression workspace is governed by $b_s$ rather than prompt length. {\color{black} The parameter $b_s$ therefore trades arithmetic efficiency against temporary memory.} Under grouped-query attention, per-key scores are aggregated at KV-head granularity before selection, so retained indices are shared within each KV group and the gathered $(\hat K,\hat V)$ remain in the compact KV-head layout. \subsection{Memory and complexity analysis}
\label{subsec:analysis}

TideKV changes the residency schedule rather than the causal attention computation: the prompt still passes through all $N$ layers, with an additional $O(WS)$ layer-end scoring pass. At the largest working set of layer $l$, \begin{equation}
\label{eq:mpeak}
\begin{aligned}
M_{\mathrm{live}}^{\mathrm{TideKV}}
\approx\;&
\underbrace{
2 B H_{\mathrm{kv}} S D b
}_{\text{active KV}}
+
\underbrace{
B S d_{\mathrm{model}} b
}_{\text{hidden}}
+
\underbrace{
c B C d_{\mathrm{ff}} b
}_{\text{chunk act.}} \\
&+
\underbrace{
\sum_{l'<l} 2 B H_{\mathrm{kv}} K_{l'} D b
}_{\text{compressed caches}}
+
\underbrace{
O\!\left(B H_q (W b_s + K_l)\right)b
}_{\text{compression temp}} .
\end{aligned}
\end{equation}
where $d_{\text{ff}}=O(d_{\text{model}})$ is the MLP intermediate width, $c$ counts co-resident MLP activation buffers, and the expression assumes a FlashAttention backend. {\color{black} Only the compressed-cache sum accumulates across layers. The active-layer KV and hidden bridge each remain linear in $S$, with footprints of $2BH_{\mathrm{kv}}SDb$ and $BSd_{\mathrm{model}}b$ bytes. Reusing hidden chunk slots avoids a separate bridge per layer. These remaining linear terms limit scaling; layer-specific workspace and allocation behavior also affect the measured peak.} {\color{black} A layer-end post-prefill compressor has the same active-layer KV and hidden terms. Whole-prompt execution gives an activation term of $cBSd_{\text{ff}}b$ and a dense compression workspace of $BH_qWSb$.} TideKV reduces these terms by $S/C$ and from $O(WS)$ to $O(Wb_s)$, respectively. The activation term provides most of the reduction because it scales with the MLP width and the full context in whole-prompt execution. {\color{black} Chunk size $C$ and score block size $b_s$ control these memory--latency trade-offs; Section~\ref{subsec:overhead} evaluates the chunk-size choice.} \subsection{Prompt fidelity and scope}
\label{subsec:fidelity}

Layer-major delayed eviction preserves each layer's prompt path up to compression. {\color{black} Consider deterministic causal layers with token-local normalization and MLPs, identical weights, tokens, positions and masks, and exact arithmetic. Assume that the input hidden sequence to layer $l$ matches uncompressed causal prefill. Each chunk then has the same queries and complete causal KV history, with the within-chunk mask excluding future tokens. Its output agrees with the reference, so induction over chunks gives the same complete hidden sequence $H_l$ and pre-compression KV. Layer-end compression affects only the retained decode cache, leaving $H_l$ unchanged; induction over layers completes the argument.

Finite-precision kernels and operation order can introduce numerical differences. We evaluate hidden-state relative-$L_2$ error and cosine similarity, together with next-token top-1 agreement and Jensen--Shannon divergence, across full, standard-chunked, no-eviction layer-major, and compressed TideKV paths (Section~\ref{subsec:fidelity-eval}). Selector-matched downstream scores assess the utility of the retained cache separately from this prompt-encoding property.}

\section{System implementation}
\label{sec:impl}

\begin{figure*}[pos=htbp] \centering \includegraphics[width=0.9\textwidth]{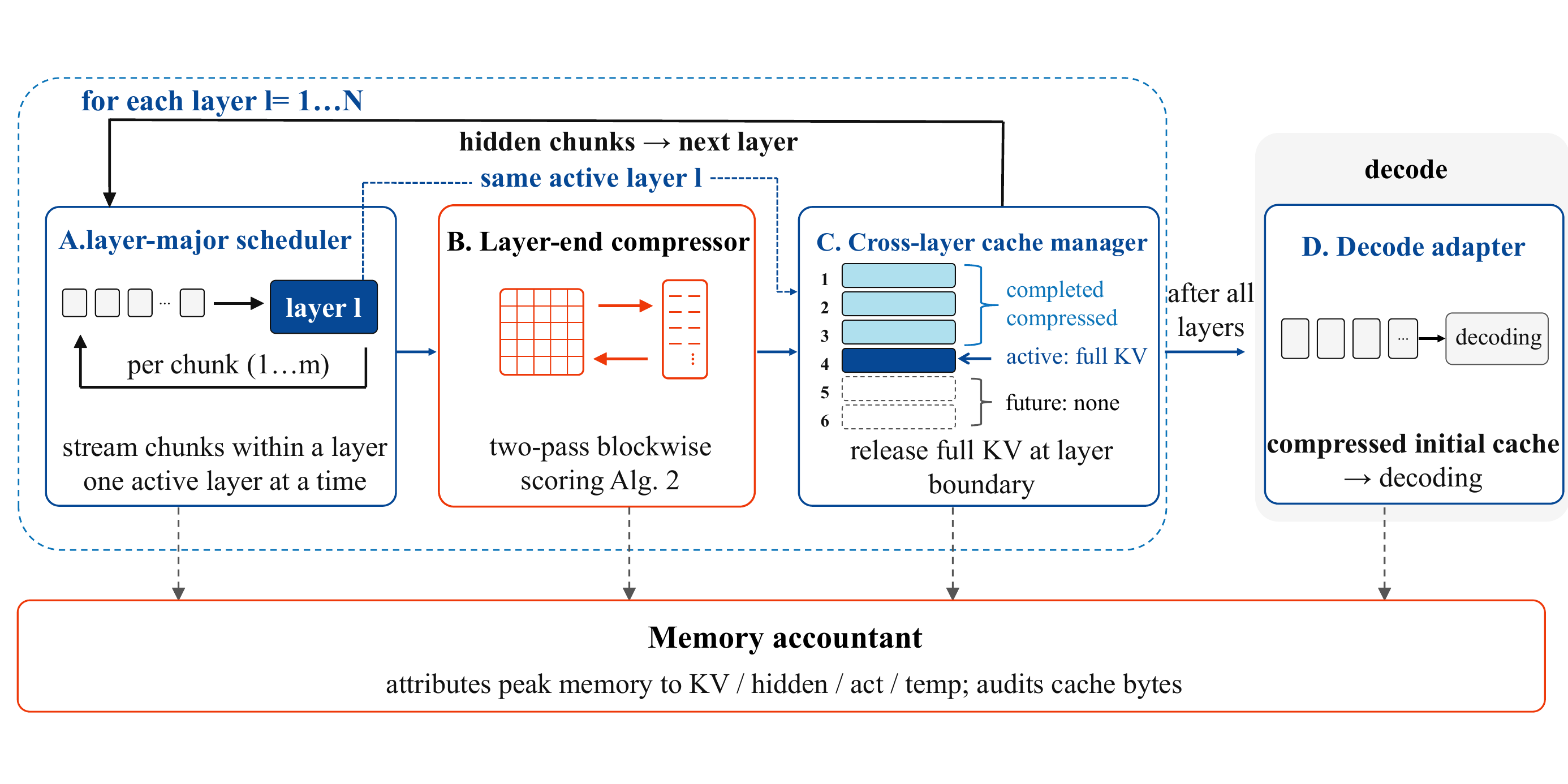} \caption{Logical architecture of the TideKV reference implementation. The layer-major scheduler coordinates the layer-end compressor and cross-layer cache manager to transform each layer's full KV into a compact retained cache at the layer boundary. After all layers complete, the decode adapter exposes the assembled initial cache to generation. The memory accountant records component residency and allocator/process-level peaks.}
\label{fig:arch}
\end{figure*}

We implement TideKV as an on-GPU prefill subsystem in Hugging Face Transformers. It reuses the pretrained weights and configured Transformer operators, but replaces the standard whole-model prompt traversal with a layer-by-chunk loop. Figure~\ref{fig:arch} summarizes the four logical components and the cross-cutting memory accountant. 

\textbf{Layer-major scheduler and hidden-state bridge.} At layer $l$, TideKV stores output chunks $H_l=\{h_1^l,\ldots,h_m^l\}$ as the hidden-state bridge to layer $l+1$ and reuses the same chunk slots as execution advances. The bridge contains $O(BS d_{\mathrm{model}})$ elements, whereas temporary attention and MLP activations remain bounded by one chunk. The active layer separately accumulates its uncompressed KV. Causal masks and absolute token positions are maintained for every chunk.

\textbf{Layer-end compression and cache management.} After the last chunk, the compressor scores the complete layer-local history, gathers the retained K/V entries and their original positions, and commits the compressed cache. The cross-layer cache manager then releases the full active-layer KV. It supports rectangular per-layer caches and a variable-length layout for head-adaptive budgets. The layer-end commit is the synchronization point before the scheduler advances to layer $l+1$.

\textbf{Decode adapter and accounting.} The decode adapter exposes the compressed initial cache in the structure expected by generation and resumes from true absolute positions rather than the shortened cache length. Throughout cache construction, the memory accountant tracks active allocation, allocator-reserved memory, final physical-cache bytes, and process-level device memory. These signals connect the component lifetimes in Fig.~\ref{fig:arch} to the end-to-end memory behavior evaluated in Section~\ref{sec:eval}.

The layer-major order changes orchestration rather than the numerical definition of attention, normalization, or MLP operators, and therefore remains compatible with intra-layer fused kernels such as FlashAttention-2. {\color{black} Its overhead includes chunk dispatch, blockwise scoring, hidden-state bridge traffic, KV gathering and packing, and layer-boundary synchronization;} these costs are included in the measured TTFT and quantified in Section~\ref{subsec:overhead}.

{\color{black} The components interact at each layer boundary: after the complete prompt is processed, the scheduler triggers compression, the cache manager commits retained KV and releases the full active-layer KV, and the next layer receives the hidden-state bridge. The memory accountant tracks these lifetimes. This coordination preserves full-prompt selection while limiting uncompressed KV residency to one active layer.} 

\Needspace{5\baselineskip}
\section{Evaluation}
\label{sec:eval}
{\color{black} We evaluate cache utility, memory savings, runtime and device robustness, and latency through selector-matched measurements, allocator/backend and persistent-request tests, cross-device experiments, and fidelity and timing analyses.} \subsection{Evaluation setup}
\label{subsec:setup}

\textbf{Models and platforms.} We evaluate Llama-3.1-8B-Instruct and Mistral-7B-Instruct-v0.2 in BF16 on an NVIDIA A100 PCIe 40-GB GPU, a 24-GB GeForce RTX 4090, and a 64-GB Jetson AGX Orin in MODE-50W. Unless stated otherwise, batch size is one, the chunk size is 2,048, the physical KV budget is 4,096 tokens, and the main attention path uses FlashAttention-2. Table~\ref{tab:experimental-setup} summarizes the principal configuration; every matched pair uses the same software stack.

\begin{table}[pos=t]
\centering
\footnotesize
\caption{Principal experimental configuration. Platform-specific software is held fixed within every matched comparison.}
\label{tab:experimental-setup}
\begin{tabularx}{\columnwidth}{@{}p{0.22\columnwidth}X@{}}
\toprule
Item & Configuration \\
\midrule
Models & Llama-3.1-8B-Instruct; Mistral-7B-Instruct-v0.2; BF16 \\
Accelerators & A100 PCIe 40~GB; RTX 4090 24~GB; Jetson AGX Orin 64~GB (MODE\_50W) \\
A100 stack & CUDA 11.8; PyTorch 2.3.1; Transformers 4.44.2; FlashAttention 2.7.4.post1 \\
Inference & Batch 1; chunk 2,048; physical KV budget 4,096 unless stated otherwise \\
\bottomrule
\end{tabularx}
\end{table}


\begin{table*}[pos=tp]
    \centering
    \caption{Detailed results of Llama-3.1-8B-Instruct and Mistral-7B-Instruct-v0.2 on LongBench. {\color{black} Bold scores indicate the higher average within each selector pair.}}
    \label{tab:longbench}
    \resizebox{\textwidth}{!}{
    \begin{tabular}{l ccc ccc ccc ccc cc cc | c}
        \toprule
        & \multicolumn{3}{c}{Single-Doc. QA}
        & \multicolumn{3}{c}{Multi-Doc. QA}
        & \multicolumn{3}{c}{Summarization}
        & \multicolumn{3}{c}{Few-shot Learning}
        & \multicolumn{2}{c}{Synthetic}
        & \multicolumn{2}{c}{Code}
        & \\
        \cmidrule(lr){2-4} \cmidrule(lr){5-7} \cmidrule(lr){8-10} \cmidrule(lr){11-13} \cmidrule(lr){14-15} \cmidrule(lr){16-17}
        & \rotatebox{-45}{NrtvQA} & \rotatebox{-45}{Qasper} & \rotatebox{-45}{MF-en} & \rotatebox{-45}{HotpotQA} & \rotatebox{-45}{2WikiMQA} & \rotatebox{-45}{Musique} & \rotatebox{-45}{GovReport} & \rotatebox{-45}{QMSum} & \rotatebox{-45}{MultiNews} & \rotatebox{-45}{TREC} & \rotatebox{-45}{TriviaQA} & \rotatebox{-45}{SAMSum} & \rotatebox{-45}{PCount} & \rotatebox{-45}{PRe} & \rotatebox{-45}{Lcc} & \rotatebox{-45}{RB-P} & \begin{tabular}[c]{@{}c@{}}Ave.\\Score\end{tabular} \\
        \midrule
        Full Cache
        & 29.69 & 44.92 & 55.78
        & 54.68 & 47.26 & 30.89
        & 35.09 & 23.55 & 27.23
        & 73.0 & 90.23 & 44.22
        & 10.38  & 99.5
        & 64.58 & 58.97
        & 49.37 \\
        \multicolumn{18}{c}{Llama-3.1-8B-Instruct,$Budget = 256$} \\
        \midrule
        snap & 26.9 & 31.93 & 52.54 & 54.55 & 43.2 & 28.41 & 22.19 & 23.81 & 22.22 & 55.6 & 85.27 & 40.94 & 6.75 & 99 & 61.69 & 54.93 & 44.37 \\
        \textbf{tidekv-snap} & 26.83 & 32.99 & 53.25 & 54.99 & 43.6 & 28.72 & 21.91 & 23.75 & 22.55 & 55.5 & 86.72 & 41.3 & 5.28 & 99.5 & 62.32 & 55.67 & \textbf{44.68} \\
        pyramidkv & 29.17 & 36.12 & 52.84 & 54.72 & 43.8 & 29.87 & 23.13 & 22.87 & 22.6 & 54.5 & 76.42 & 41.64 & 6.78 & 99.5 & 60.54 & 51.97 & \textbf{44.15} \\
        \textbf{tidekv-pyramidkv} & 29.31 & 35.86 & 53.86 & 53.9 & 43.32 & 30.19 & 23.08 & 23.27 & 22.28 & 54 & 76.77 & 40.91 & 5.28 & 99 & 60.66 & 52.66 & 44.02 \\
        adakv & 26.05 & 32.78 & 52.78 & 55.77 & 43.46 & 27.55 & 22.62 & 23.43 & 22.67 & 58.5 & 86.87 & 40.91 & 6.78 & 99.5 & 62.56 & 57.09 & 44.96 \\
        \textbf{tidekv-adakv} & 25.22 & 33.98 & 54.74 & 54.98 & 42.87 & 29.37 & 21.97 & 24.04 & 22.4 & 59.5 & 87.86 & 41.28 & 5.28 & 99.5 & 62.36 & 56.47 & \textbf{45.11 }\\
        chunkkv & 25.61 & 28.66 & 52.28 & 53.78 & 43.63 & 27.46 & 20.34 & 22.9 & 21.39 & 47.5 & 89.54 & 41.27 & 5.69 & 99 & 61.14 & 54.65 & 43.43 \\
        \textbf{tidekv-chunkkv} & 25.25 & 30.22 & 53.6 & 54.22 & 41.74 & 29.47 & 20.37 & 23.37 & 21.82 & 47 & 88.16 & 41.45 & 5.36 & 98 & 62.6 & 54.82 & \textbf{43.59} \\
        \midrule
        \multicolumn{18}{c}{Llama-3.1-8B-Instruct,$Budget =1024$} \\
        \midrule
        snap & 29.4 & 42.61 & 53.43 & 55.79 & 44.78 & 29.16 & 27.81 & 24.37 & 25.99 & 67.5 & 89.62 & 42.91 & 6.83 & 99.5 & 64.25 & 58.01 & 47.62 \\
        \textbf{tidekv-snap} & 29.51 & 43.27 & 54.65 & 55.49 & 44.24 & 31.4 & 27.11 & 24.61 & 25.57 & 69 & 90.61 & 42.23 & 5.25 & 99.5 & 64.19 & 58.18 & \textbf{47.8} \\
        pyramidkv & 29.32 & 43.27 & 55.26 & 55.56 & 44.08 & 31.39 & 28.53 & 23.92 & 25.9 & 70.5 & 90.17 & 42.13 & 6.75 & 99 & 63.25 & 57 & 47.88 \\
        \textbf{tidekv-pyramidkv} & 29.25 & 43.48 & 55.87 & 55.22 & 44.05 & 31.65 & 28.01 & 24.14 & 25.88 & 70.5 & 90.98 & 43.67 & 5.25 & 99.5 & 64.02 & 55.97 & \textbf{47.96} \\
        adakv & 30.13 & 43.33 & 55.25 & 56.17 & 44.75 & 29.54 & 27.58 & 24.57 & 25.89 & 68.5 & 90.37 & 42.03 & 6.75 & 99.5 & 64.04 & 58.43 & \textbf{47.93} \\
        \textbf{tidekv-adakv} & 30.31 & 43.1 & 55.31 & 55.63 & 44.36 & 30.62 & 27.43 & 24.94 & 25.66 & 70 & 89.94 & 42.35 & 5.25 & 99.5 & 64.07 & 58.3 & 47.92 \\
        chunkkv & 28.65 & 43.66 & 54.54 & 56.35 & 45.61 & 31.54 & 26.84 & 23.91 & 25.66 & 67 & 90.41 & 42.29 & 6.78 & 98.5 & 64.55 & 58.24 & 47.78 \\
        \textbf{tidekv-chunkkv} & 28.63 & 43.77 & 55.81 & 55.45 & 44.45 & 31.02 & 27.04 & 24.46 & 25.67 & 68.5 & 91 & 42.85 & 5.28 & 99.5 & 65.08 & 58.55 & \textbf{47.94} \\
        \midrule
        \multicolumn{18}{c}{Llama-3.1-8B-Instruct,$Budget=4096$} \\
        \midrule
        snap & 29.72 & 44.45 & 55.3 & 56.51 & 45.14 & 29.89 & 32.27 & 25.28 & 26.95 & 71.5 & 91.48 & 43.62 & 6.75 & 99.5 & 64.61 & 58.95 & 48.87 \\
        \textbf{tidekv-snap} & 30.2 & 44.64 & 56.4 & 55.34 & 44.63 & 31.65 & 32.29 & 24.96 & 27.12 & 72.5 & 90.98 & 43.41 & 5.25 & 99.5 & 64.81 & 58.66 & \textbf{48.9} \\
        pyramidkv & 30.4 & 44.25 & 55.43 & 56.47 & 45.14 & 29.88 & 31.55 & 24.53 & 27.03 & 72.5 & 91.65 & 43.77 & 6.75 & 99.5 & 64.49 & 58.9 & \textbf{48.89} \\
        \textbf{tidekv-pyramidkv} & 29.59 & 44.64 & 56.5 & 55.39 & 45.14 & 31.03 & 32.57 & 24.83 & 27.05 & 72.5 & 90.98 & 43.62 & 5.25 & 99.5 & 64.59 & 58.45 & 48.85 \\
        adakv & 30.16 & 45.04 & 55.5 & 56.23 & 45.04 & 29.5 & 32.38 & 25.41 & 26.79 & 71.5 & 91.48 & 43.59 & 6.75 & 99.5 & 64.68 & 59.11 & 48.92 \\
        \textbf{tidekv-adakv} & 30.32 & 44.68 & 56.03 & 55.53 & 44.98 & 31.2 & 32.6 & 25.22 & 26.99 & 72.5 & 90.98 & 43.55 & 5.25 & 99.5 & 64.73 & 59.19 & \textbf{48.95} \\
        chunkkv & 29.13 & 44.43 & 55.56 & 56.07 & 44.65 & 29.77 & 32.3 & 24.36 & 26.93 & 72.5 & 91.65 & 43.04 & 6.75 & 99.5 & 64.52 & 58.35 & 48.72 \\
        \textbf{tidekv-chunkkv} & 29.83 & 44.81 & 56.37 & 55.1 & 44.59 & 31.31 & 32.47 & 25.7 & 27.05 & 72.5 & 90.68 & 43.4 & 5.25 & 99.5 & 64.86 & 59.25 & \textbf{48.92} \\

        \midrule
        \multicolumn{18}{c}{Mistral-7B-Instruct-v0.2, $Budget=256$} \\
        \midrule
        snap & 17.74 & 21.85 & 43.26 & 26.6 & 16.48 & 12.43 & 22.16 & 22.57 & 22.92 & 57.5 & 85.79 & 40.05 & 2.96 & 76.82 & 54.28 & 51.67 & \textbf{35.94} \\
        \textbf{tidekv-snap} & 18.32 & 20.33 & 42.42 & 26.89 & 16.27 & 11.93 & 22.44 & 22.1 & 22.55 & 57 & 85.51 & 39.49 & 2.95 & 76.33 & 54.09 & 50.95 & 35.6 \\
        pyramidkv & 18.01 & 19.64 & 41.65 & 27.52 & 15.38 & 12.12 & 22.75 & 21.71 & 22.38 & 68 & 84.81 & 40.31 & 2.36 & 77.44 & 52.68 & 49.23 & 36 \\
        \textbf{tidekv-pyramidkv} & 17.97 & 21.26 & 41.89 & 30.28 & 15.97 & 12.44 & 22.76 & 22.03 & 22.08 & 66.5 & 84.22 & 39.72 & 1.96 & 77.73 & 52.98 & 47.29 & \textbf{36.07} \\
        chunkkv & 16.98 & 19.97 & 44.88 & 27.05 & 16.6 & 14 & 22.2 & 22.08 & 21.96 & 65 & 86.35 & 38.92 & 3.22 & 53.69 & 53.58 & 48.29 & \textbf{34.67} \\
        \textbf{tidekv-chunkkv} & 18.61 & 20.05 & 45.47 & 27.51 & 15.95 & 13.29 & 20.47 & 21.29 & 21.64 & 64.5 & 85.19 & 39.47 & 4.06 & 53.06 & 54.24 & 49.53 & 34.64 \\
        \midrule
        \multicolumn{18}{c}{Mistral-7B-Instruct-v0.2,$Budget=1024$} \\
        \midrule
        snap & 20.3 & 27.75 & 45.78 & 32.72 & 19.04 & 16.74 & 26.87 & 23.82 & 25.74 & 68 & 86.34 & 40.93 & 3.29 & 88.56 & 57.59 & 53.41 & \textbf{39.8} \\
        \textbf{tidekv-snap} & 20.33 & 26.84 & 45.95 & 33.54 & 18.97 & 16.59 & 26.7 & 23.5 & 25.71 & 67 & 86.28 & 41.37 & 3.08 & 87.29 & 56.91 & 53.67 & 39.61 \\
        pyramidkv & 20.39 & 27.13 & 45.21 & 32.34 & 19.01 & 15.42 & 26.4 & 23 & 25.64 & 69.5 & 85.96 & 41.28 & 4.36 & 86.08 & 55.67 & 52.57 & \textbf{39.37} \\
        \textbf{tidekv-pyramidkv} & 19.78 & 27.55 & 45.23 & 32.32 & 19.94 & 14.73 & 26.5 & 22.83 & 25.49 & 68 & 85.64 & 41.64 & 3.6 & 87.33 & 56.07 & 52.25 & 39.31 \\
        chunkkv & 19.02 & 26.76 & 45.94 & 32.39 & 18.75 & 14.81 & 25.95 & 21.5 & 25.44 & 66 & 86.34 & 40.82 & 3.53 & 82.6 & 57.14 & 53.61 & 38.79 \\
        \textbf{tidekv-chunkkv} & 20.13 & 27.86 & 46.56 & 33.98 & 19.19 & 16.67 & 26.78 & 23.39 & 25.79 & 68 & 86.34 & 40.37 & 2.72 & 85.92 & 57.08 & 53.2 & \textbf{39.62} \\
        \midrule
        \multicolumn{18}{c}{Mistral-7B-Instruct-v0.2,$Budget=4096$} \\
        \midrule
        snap & 21.6 & 29.27 & 47.03 & 35.64 & 21.78 & 19 & 31.2 & 23.67 & 26.87 & 70 & 86.24 & 41.74 & 4.04 & 89.17 & 57.23 & 54.4 & \textbf{41.18} \\
        \textbf{tidekv-snap} & 21.68 & 30.02 & 46.91 & 35.6 & 21.84 & 19.42 & 31.31 & 23.68 & 26.84 & 70.5 & 86.23 & 41.82 & 3.41 & 87.96 & 57.2 & 54.27 & 41.17 \\
        pyramidkv & 21.34 & 29.28 & 46.28 & 34.54 & 21.66 & 18.66 & 30.86 & 23.89 & 26.78 & 71 & 86.17 & 41.89 & 3.11 & 88.21 & 57.41 & 54.45 & \textbf{40.97} \\
        \textbf{tidekv-pyramidkv} & 21.18 & 29.38 & 46.18 & 34.18 & 21.02 & 19.1 & 30.5 & 23.94 & 26.92 & 71 & 86.28 & 41.68 & 3.66 & 87.12 & 57.25 & 53.64 & 40.81 \\
        chunkkv & 21.08 & 29.33 & 46.99 & 36.05 & 21.16 & 18.86 & 30.96 & 23.27 & 26.87 & 71 & 86.24 & 41.85 & 2.79 & 87.94 & 57.37 & 54.8 & 41.04 \\
        \textbf{tidekv-chunkkv} & 21.63 & 29.31 & 46.98 & 35.82 & 21.27 & 19.39 & 31.5 & 23.76 & 26.84 & 71 & 86.25 & 42.23 & 3.05 & 88.06 & 57.31 & 54.11 & \textbf{41.15} \\
        \bottomrule
    \end{tabular}
    }
\end{table*}

\textbf{Benchmarks and metrics.} LongBench~\cite{bai2024longbench} contains 16 English tasks and 3,750 examples. RULER~\cite{hsieh2024ruler} contains 13 controlled retrieval and aggregation tasks; we use 100 examples per task at 4K--128K and report both mean score and completion status. Memory measurements distinguish peak allocated, peak reserved, and process-level NVML memory on discrete GPUs. {\color{black} On Jetson, we report CUDA-managed allocation because the CPU and GPU share physical memory.} Time to first token (TTFT) includes prefill, cache selection, and the first generated token.

\textbf{Baselines and matched controls.} The baselines span token selection (SnapKV, PyramidKV, AdaKV, and ChunkKV), online eviction (KeyDiff~\cite{park2026keydiff}), KV quantization (KIVI~\cite{liu2024kivi}), and CPU--GPU head offloading (HeadInfer~\cite{luo2025headinfer}). Each schedule-level comparison holds model weights, prompt tokens, precision, selector parameters, physical KV budget, decode settings, attention backend, allocator, synchronization, and measurement procedure fixed. The detailed construction analysis uses a 4,096-token budget, which produces a 512-MiB compact GQA cache for both methods.

\subsection{Selector-matched quality and memory}
\label{subsec:overall}

We first examine whether TideKV lowers the memory required to construct a cache without changing its utility or final size. Each selector is paired with its TideKV counterpart at physical KV budgets of 256, 1,024, and 4,096 tokens and evaluated on LongBench and RULER.

\begin{table*}[pos=!t]
\centering
\fontsize{8pt}{9.6pt}\selectfont
\setlength{\tabcolsep}{1.3pt}
\renewcommand{\arraystretch}{1.12}
\caption{{\color{black} RULER performance (\%) of Llama-3.1-8B-Instruct on A100 under different physical KV budgets and context lengths. Avg equally weights the five displayed lengths. SnapKV and PyramidKV denote selector-matched controls. Higher is better.}}
\label{tab:ruler_llama}
\begin{tabular*}{\textwidth}{@{\extracolsep{\fill}}l|rrrrrr|rrrrrr|rrrrrr@{}}
\toprule
\multirow{2}{*}{Method} & \multicolumn{6}{c|}{Budget = 256} & \multicolumn{6}{c|}{Budget = 1024} & \multicolumn{6}{c}{Budget = 4096} \\
\cmidrule(lr){2-7}\cmidrule(lr){8-13}\cmidrule(lr){14-19}
 & 4K & 16K & 32K & {\color{black} 64K} & {\color{black} 128K} & {\color{black} Avg} & 4K & 16K & 32K & {\color{black} 64K} & {\color{black} 128K} & {\color{black} Avg} & 4K & 16K & 32K & {\color{black} 64K} & {\color{black} 128K} & {\color{black} Avg} \\
\midrule
SnapKV & 68.47 & 65.68 & 62.97 & {\color{black} 60.20} & {\color{black} 43.31} & {\color{black} 60.13} & 88.56 & 78.78 & 74.46 & {\color{black} 68.96} & {\color{black} 58.92} & {\color{black} 73.94} & 95.18 & 89.50 & 81.12 & {\color{black} 76.72} & {\color{black} 65.90} & {\color{black} 81.68} \\
\textbf{TideKV-Snap} & 69.70 & 67.88 & 64.62 & {\color{black} 59.95} & {\color{black} 42.94} & {\color{black} 61.02} & 88.25 & 78.57 & 74.28 & {\color{black} 69.14} & {\color{black} 59.07} & {\color{black} 73.86} & 95.00 & 90.47 & 82.07 & {\color{black} 78.07} & {\color{black} 68.51} & {\color{black} 82.82} \\
\midrule
PyramidKV & 61.64 & 61.35 & 58.61 & {\color{black} 53.93} & {\color{black} 39.24} & {\color{black} 54.95} & 80.58 & 73.66 & 70.97 & {\color{black} 65.72} & {\color{black} 56.09} & {\color{black} 69.40} & 95.18 & 87.71 & 77.34 & {\color{black} 71.73} & {\color{black} 63.29} & {\color{black} 79.05} \\
\textbf{TideKV-PyramidKV} & 61.70 & 61.11 & 59.48 & {\color{black} 54.29} & {\color{black} 40.68} & {\color{black} 55.45} & 80.21 & 73.50 & 70.09 & {\color{black} 65.60} & {\color{black} 56.17} & {\color{black} 69.11} & 95.00 & 85.30 & 77.66 & {\color{black} 71.66} & {\color{black} 63.35} & {\color{black} 78.59} \\
\midrule
AdaKV & 70.05 & 68.21 & 63.98 & {\color{black} 59.38} & {\color{black} 41.43} & {\color{black} 60.61} & 88.50 & 79.90 & 75.42 & {\color{black} 68.85} & {\color{black} 58.46} & {\color{black} 74.23} & 95.09 & 90.05 & 81.93 & {\color{black} 77.25} & {\color{black} 65.70} & {\color{black} 82.00} \\
\textbf{TideKV-AdaKV} & 70.20 & 68.10 & 64.14 & {\color{black} 59.22} & {\color{black} 41.12} & {\color{black} 60.56} & 89.80 & 79.67 & 75.02 & {\color{black} 69.10} & {\color{black} 58.24} & {\color{black} 74.37} & 95.00 & 90.74 & 83.31 & {\color{black} 77.87} & {\color{black} 68.19} & {\color{black} 83.02} \\
\midrule
ChunkKV & 73.23 & 73.57 & 72.32 & {\color{black} 66.68} & {\color{black} 34.97} & {\color{black} 64.15} & 91.53 & 84.78 & 81.88 & {\color{black} 80.14} & {\color{black} 65.62} & {\color{black} 80.79} & 95.18 & 89.22 & 85.85 & {\color{black} 82.35} & {\color{black} 73.11} & {\color{black} 85.14} \\
\textbf{TideKV-ChunkKV} & 73.59 & 73.23 & 72.41 & {\color{black} 66.27} & {\color{black} 34.63} & {\color{black} 64.03} & 91.83 & 84.25 & 82.57 & {\color{black} 80.07} & {\color{black} 66.07} & {\color{black} 80.96} & 95.00 & 89.56 & 84.66 & {\color{black} 83.35} & {\color{black} 75.14} & {\color{black} 85.54} \\
\bottomrule
\end{tabular*}
\end{table*}

Across both benchmarks, the paired variants remain close. Table~\ref{tab:longbench} shows that TideKV preserves the selector-specific LongBench quality profile. {\color{black} Table~\ref{tab:ruler_llama} reports the corresponding A100 RULER results from 4K to 128K. Across the 12 selector--budget pairs, the five-length mean score changes by $-0.46$ to $+1.14$ points under TideKV. Differences at individual lengths range from $-2.41$ to $+2.61$ points. At 64K and 128K, all four selector families remain close to their matched baselines, with differences of $-0.41$ to $+2.61$ points across 24 comparisons. At budget 1,024, the mean absolute difference over these two lengths is 0.19 points. Thus, the agreement between each selector and its TideKV counterpart persists across both cache budgets and context lengths.}

\begin{figure*}[pos=t]
\centering
    \includegraphics[width=\textwidth]{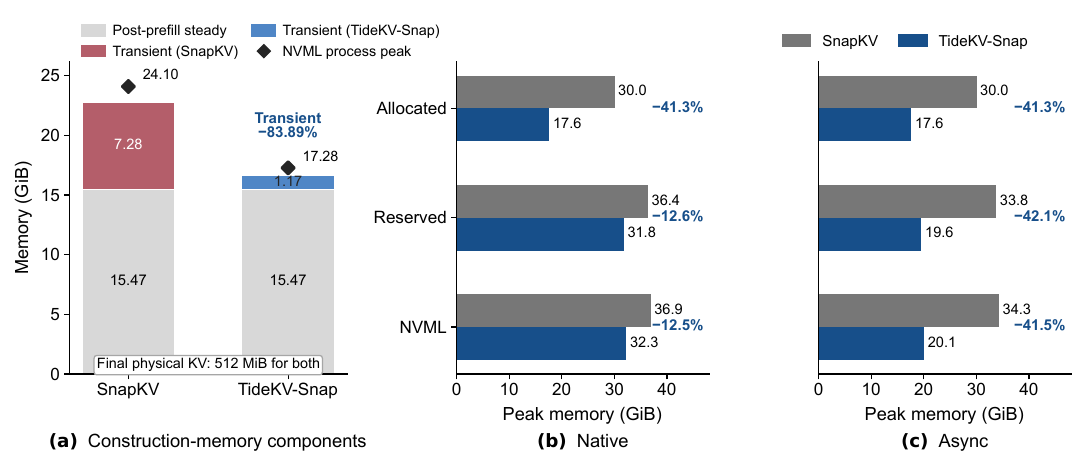}
\caption{Matched construction-memory accounting on A100. (a) At 64K, both methods reach the same 15.466-GiB post-prefill allocation and retain the same 512-MiB physical cache; TideKV-Snap reduces the transient construction footprint from 7.281 to 1.173~GiB. Diamonds mark the NVML process peaks. (b,c) At 128K, the allocated-peak reduction is identical under Native and Async, while allocator retention changes the reserved and process-level realization. Percentages denote reductions from SnapKV to TideKV-Snap.}
\label{fig:memory-accounting}
\end{figure*}

The matched memory accounting in Fig.~\ref{fig:memory-accounting} identifies where the saving occurs. After prefill, both methods converge to the same 15.466-GiB steady allocation, which comprises the loaded model, the identical 512-MiB compact cache, and the remaining persistent runtime tensors; it is not the model-weight footprint alone. During cache construction, TideKV-Snap lowers peak allocation from 22.748 to 16.639~GiB and the NVML process peak from 24.096 to 17.283~GiB. Subtracting the common steady state isolates a transient footprint of 7.281~GiB for SnapKV and 1.173~GiB for TideKV-Snap, an 83.89\% reduction. TideKV therefore reduces the temporary memory required to build the same final cache rather than shrinking the cache used for decoding.

\subsection{Runtime-memory robustness}
\label{subsec:robustness}

We next examine whether the active-memory saving depends on the allocator or attention implementation, and whether repeated mixed-length requests cause allocator growth. The allocator sweep uses the A100, Llama-3.1-8B-Instruct in BF16, batch size one, a 4,096-token physical budget, $C=2048$, and FlashAttention-2. ``Native'' denotes PyTorch's default caching allocator; ``Async'' denotes the CUDA stream-ordered \texttt{cudaMallocAsync} backend selected through PyTorch's allocator configuration. The method pair and all other settings are unchanged between these runs.

\begin{table*}[pos=t]
\centering
\scriptsize
\caption{Backend and persistent-workload robustness on A100. Positive values in panel (a) denote reductions from SnapKV to TideKV-Snap; negative values denote increases. Panel (b) reports 150 requests per configuration (600 total).}
\label{tab:runtime-robustness}
\begin{minipage}[t]{0.43\textwidth}
\vspace{0pt}
\centering
\textbf{(a) Attention-backend stress tests}\par\smallskip
\setlength{\tabcolsep}{2.1pt}
\begin{tabular}{@{}llrrr@{}}
\toprule
Backend & Context & Alloc. & Reserved & NVML \\
\midrule
FA2 & 8K  & 3.74\% & 5.41\% & 5.26\% \\
FA2 & 16K & 8.08\% & 11.08\% & 10.79\% \\
FA2 & 32K & 15.53\% & 21.38\% & 20.89\% \\
FA2 & 64K & 26.85\% & 24.59\% & 24.13\% \\
SDPA & 8K  & 3.55\% & 4.81\% & 4.68\% \\
SDPA & 16K & 7.04\% & 9.97\% & 9.70\% \\
SDPA & 32K & 12.96\% & 6.70\% & 6.54\% \\
SDPA & 64K & 21.94\% & $-23.39$\% & $-22.94$\% \\
\bottomrule
\end{tabular}
\end{minipage}\hfill
\begin{minipage}[t]{0.55\textwidth}
\vspace{0pt}
\centering
\textbf{(b) Persistent mixed-length traces}\par\smallskip
\setlength{\tabcolsep}{1.5pt}
\resizebox{\linewidth}{!}{%
\begin{tabular}{@{}llrrrrr@{}}
\toprule
Allocator & Method & Completed & Max reserved & Slope & Cycle $\Delta$ & Inactive / retries \\
\midrule
Async & SnapKV & 150 & 33.781 GiB & $\approx0$ & 0 MiB & -- / -- \\
Async & TideKV-Snap & 150 & 19.625 GiB & $0$ & 0 MiB & -- / -- \\
Native & SnapKV & 150 & 36.828 GiB & $\approx0$ & 0 MiB & 7.234 GiB / 1 \\
Native & TideKV-Snap & 150 & 32.154 GiB & $0$ & 0 MiB & 1.326 GiB / 0 \\
\bottomrule
\end{tabular}%
}
\end{minipage}
\end{table*}

\textbf{Allocator sensitivity.} At 128K, Fig.~\ref{fig:memory-accounting}(b,c) shows the same 41.26\% allocated-peak reduction under Native and Async, while reserved and NVML reductions range from 12.47\% to 42.09\%. Thus, changing the allocator leaves the active-memory reduction unchanged but alters how much of it appears at the process level.

\textbf{Attention-backend sensitivity.} The backend stress tests in Table~\ref{tab:runtime-robustness}(a) support the same distinction. Under FlashAttention-2, allocated-peak savings increase from 3.74\% at 8K to 26.85\% at 64K, while reserved and NVML peaks also decrease. With SDPA, allocated-peak savings increase from 3.55\% to 21.94\% over the same range. At 64K, however, SDPA raises the reserved and NVML peaks by 23.39\% and 22.94\%, respectively, showing that backend workspaces can mask lower active residency at process level. Eager attention further exposes the feasibility boundary: SnapKV is OOM at both 8K and 16K, whereas TideKV-Snap completes both contexts.

\textbf{Persistent-workload stability.} To test whether allocator history causes memory to accumulate across sequential requests, we executed a 150-request mixed-length trace for each allocator--method configuration in a persistent process. Stability requires the post-warm-up reserved-memory trajectory to reach a plateau rather than grow across repeated cycles. As shown in Table~\ref{tab:runtime-robustness} (b), all 600 requests complete, the fitted reserved-memory slopes are approximately zero after warm-up, and the first-to-last cycle change is 0~MiB. These time-domain measures show that the reserved-memory peaks do not accumulate with request history. With Native, TideKV-Snap further reduces maximum inactive split memory from 7.234 to 1.326~GiB (81.67\%) and reduces allocation retries from one to zero. Thus, the lower construction footprint remains stable under the tested persistent single-request workload while also reducing native-allocator fragmentation.

\subsection{Device feasibility and long-context quality}
\label{subsec:feasibility}

We then measure whether the lower construction peak extends feasible prompt length on memory-constrained hardware while retaining long-context accuracy. The first-token capacity sweep measures prefill feasibility on the RTX 4090 and Jetson; the independent 32-token protocol is analyzed with the latency results in Section~\ref{subsec:overhead}.

\textbf{Cross-device capacity.}

\begin{table}[pos=!t]
\centering
\footnotesize
\setlength{\tabcolsep}{4pt}
\caption{First-token capacity on resource-constrained platforms. Entries are peak memory in GiB: NVML process memory on the RTX 4090 and CUDA-managed allocation on Jetson. OOM denotes prefill failure. Every completed method retains a 512-MiB final physical KV cache.}
\label{tab:device-capacity}
\begin{tabular}{@{}lccc@{}}
\toprule
Platform & Context & SnapKV peak & TideKV-Snap peak \\
\midrule
RTX 4090 & 32K  & 20.027 & 16.771 \\
RTX 4090 & 64K  & 23.447 & 17.240 \\
RTX 4090 & 128K & OOM & 18.529 \\
\midrule
Jetson AGX Orin & 32K  & 19.099 & 16.131 \\
Jetson AGX Orin & 64K  & OOM & 16.631 \\
Jetson AGX Orin & 128K & OOM & 17.632 \\
\bottomrule
\end{tabular}
\end{table}

On the RTX 4090, TideKV-Snap reduces the 64K process peak from 23.447 to 17.240~GiB (26.47\%) and completes 128K at 18.529~GiB, whereas SnapKV is OOM (Table~\ref{tab:device-capacity}). {\color{black} At 32K, the matched capacity measurement gives a 15.53\% allocated-peak reduction and matches the independent 32-token protocol. For batch-one BF16 Llama-3.1-8B, the hidden bridge occupies 0.25, 0.50 and 1.00~GiB at 32K, 64K and 128K; active-layer KV adds 0.50~GiB at 128K. Doubling context from 64K to 128K increases the measured process peak by 1.289~GiB. The bridge and active KV remain linear in prompt length, but their storage is manageable through 128K in the tested setting. Section~\ref{subsec:overhead} quantifies the associated latency.}

On Jetson, TideKV-Snap reduces CUDA-managed allocation from 19.099 to 16.131~GiB at 32K (15.54\%). It remains feasible at 64K and 128K, with peaks of 16.631 and 17.632~GiB, respectively, whereas SnapKV is OOM at both lengths (Table~\ref{tab:device-capacity}). TideKV-Snap therefore extends the feasible context boundary from 32K to 128K on the tested unified-memory platform.

\begin{table}[pos=!t]
\centering
\scriptsize
\setlength{\tabcolsep}{2.0pt}
\caption{{\color{black} LongBench averages and RULER scores of Llama-3.1-8B-Instruct on RTX 4090. Paired RULER columns indicate context lengths. The 64K matched SnapKV score uses completed examples only.}}
\label{tab:ruler-extended}
\begin{tabular}{@{}lcccc@{}}
\toprule
Method & LongBench & 4K/8K & 16K/32K & 64K/128K \\
\midrule
TideKV-Snap & 48.90 & 95.14/93.44 & 90.63/82.17 & 76.65/65.93 \\
Matched SnapKV & 48.87 & 95.03/93.60 & 89.55/80.98 & 74.82/OOM \\
KeyDiff & 47.29 & 95.06/84.05 & 73.54/64.38 & 56.32/33.07 \\
KIVI & 48.69 & 94.04/93.31 & 91.01/84.55 & OOM/OOM \\
HeadInfer & 49.08 & 95.16/94.36 & 90.89/85.93 & 84.89/OOM \\
\bottomrule
\end{tabular}
\end{table}

{\color{black} \textbf{Very-long-context quality.} Table~\ref{tab:ruler-extended} reports quality and completion outcomes on RTX 4090. TideKV-Snap completes all evaluated RULER lengths, scoring 76.65 at 64K and 65.93 at 128K. Matched SnapKV is close through 32K, partially completes the 64K evaluation, and is OOM at 128K. At the same 4,096-token budget, TideKV-PyramidKV scores 71.50/60.16 and TideKV-AdaKV scores 76.87/66.03 at 64K/128K. Available RULER logs show similar peak allocated memory: 16.495--16.510~GiB for all three variants at 64K and 17.494--17.509~GiB for Snap/Pyramid at 128K. All five logged configurations complete without OOM. These generation-run allocated peaks are distinct from the NVML measurements in Table~\ref{tab:device-capacity}.}

\textbf{Cross-paradigm comparison.} The same table compares representative resource-management paradigms: SnapKV is the matched post-prefill selector, KeyDiff performs online KV eviction, KIVI quantizes the retained cache, and HeadInfer offloads attention heads between CPU and GPU. LongBench averages range from 47.29 to 49.08, with TideKV-Snap scoring 48.90. At 128K, KeyDiff completes but scores 33.07, while KIVI and HeadInfer are OOM; HeadInfer's higher 64K score of 84.89 therefore does not extend to 128K. {\color{black} KeyDiff maintains a fixed retained-KV budget, whereas TideKV preserves the full layer-local history during prompt encoding. The observed scores illustrate the memory--quality trade-off between these construction strategies. TideKV's lower construction peak expands the feasible context length while retaining the quality profile of its underlying selector.}

\subsection{Prompt fidelity of delayed eviction}
\label{subsec:fidelity-eval}

We isolate whether layer-major execution and delayed layer-end eviction alter the prompt representation used to construct the cache. The comparison includes an unchunked reference (Full-OneShot), BF16 chunking without eviction (Full-Chunked), the layer-major scheduler without eviction (Tide-NoEvict), TideKV with delayed compression, and two immediate-eviction controls using SnapKV or recent-token retention.

\begin{table}[pos=t]
\centering
\footnotesize
\setlength{\tabcolsep}{3.5pt}
\caption{Prompt-encoding fidelity and representation drift relative to Full-OneShot (Llama-3.1-8B, $S{=}16$K, $C{=}512$, budget $1024$, 20 prompts). Columns: hidden-state relative-$L_2$ and cosine similarity; logit top-1 agreement and Jensen--Shannon (JS) divergence between the prefill next-token distributions. Top: layer-major scheduling and delayed eviction stay at the bf16 chunking floor. Bottom: with the \emph{same} SnapKV selector, immediate eviction drifts an order of magnitude more. Standard post-prefill SnapKV coincides with the Full-OneShot row, as its prefill forward is uncompressed.}
\label{tab:drift}
\begin{tabular}{@{}l cc cc@{}}
\toprule
 & \multicolumn{2}{c}{hidden} & \multicolumn{2}{c}{logit} \\
\cmidrule(lr){2-3}\cmidrule(lr){4-5}
Mode (selector, eviction) & rel-$L_2$ & cos & top-1 & JS \\
\midrule
Full-OneShot (ref.)            & $0$     & $1.000$  & $1.00$ & $0$ \\
Full-Chunked (no evict)        & $0.012$ & $0.9999$ & $0.95$ & $3\times10^{-5}$ \\
Tide-NoEvict                   & $0.012$ & $0.9999$ & $0.95$ & $3\times10^{-5}$ \\
\textbf{TideKV} (SnapKV, del.) & $0.012$ & $0.9999$ & $0.95$ & $3\times10^{-5}$ \\
\midrule
Block-SnapKV (SnapKV, imm.)    & $0.122$ & $0.990$  & $0.55$ & $0.002$ \\
Block-Recent (recent, imm.)    & $1.169$ & $0.439$  & $0.00$ & $0.219$ \\
\bottomrule
\end{tabular}
\end{table}

At 16K with $C=512$, Full-Chunked, Tide-NoEvict, and TideKV all yield hidden relative-$L_2$ 0.012, cosine similarity 0.9999, top-1 agreement 0.95, and JS divergence $3\times10^{-5}$ relative to Full-OneShot (Table~\ref{tab:drift}). These results place the layer-major schedule and delayed eviction at the numerical floor introduced by ordinary BF16 chunking.

Immediate eviction produces substantially larger drift. Block-SnapKV raises relative-$L_2$ to 0.122 and lowers top-1 agreement to 0.55; recent-token eviction lowers cosine similarity to 0.439 and top-1 agreement to zero (Table~\ref{tab:drift}). {\color{black} This lower prompt drift is consistent with the selector-matched quality results.} \begin{figure*}[pos=tp] \centering
\begin{subfigure}[b]{0.48\textwidth}
        \centering
        \includegraphics[width=\linewidth]{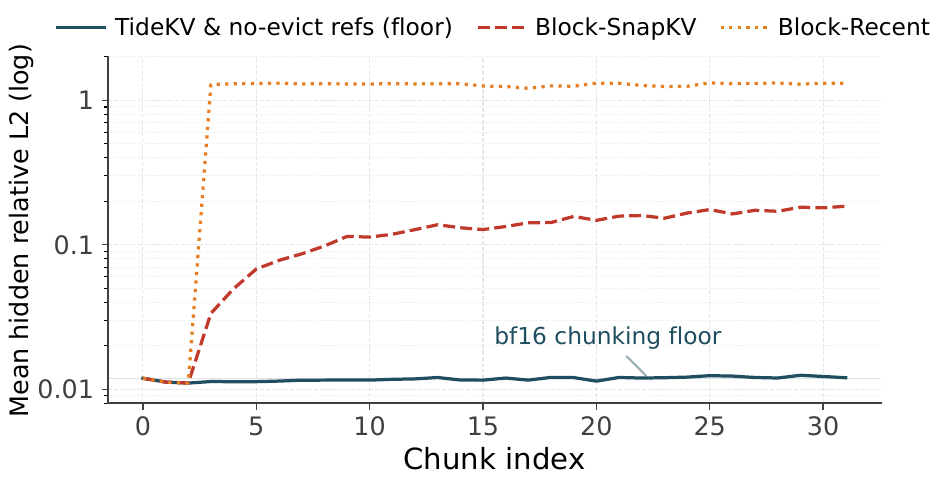}
\caption{Drift vs. chunk index.}
    \end{subfigure}\hfill
\begin{subfigure}[b]{0.48\textwidth}
        \centering
        \includegraphics[width=\linewidth]{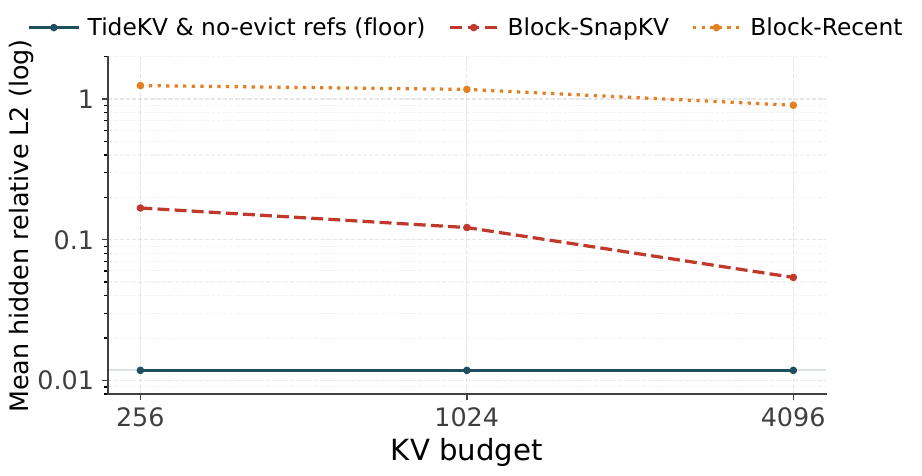}
\caption{Drift vs. KV budget.}
\end{subfigure}
\caption{Representation drift relative to Full-OneShot. Immediate Block-SnapKV drift grows with chunk index and decreases with a larger budget, whereas TideKV remains at the no-eviction chunking floor.}
\label{fig:exp-drift}
\end{figure*}

\subsection{Latency and memory--time trade-off}
\label{subsec:overhead}

Finally, we quantify the latency cost of lowering the construction peak, verify decoding with the resulting same-size cache, and identify a practical chunk-size operating point. Each latency run generates 32 tokens. {\color{black} We report TTFT, peak memory, and subsequent-token throughput.}

\begin{table*}[pos=!t]
\centering
\footnotesize
\setlength{\tabcolsep}{3.2pt}
\caption{Matched 32-token latency and memory results. Values are medians; memory is the NVML process peak on A100/RTX 4090 and CUDA-managed allocation on Jetson. {\color{black} Each completed method starts decoding from a 512-MiB compressed KV cache.}}
\label{tab:latency-boundary}
\begin{tabular}{@{}llcccc@{}}
\toprule
Platform & Context & SnapKV TTFT / peak & TideKV-Snap TTFT / peak & TTFT change & Peak change \\
\midrule
A100 & 64K & 12.390 s / 24.096 GiB & 14.464 s / 17.283 GiB & +16.74\% & $-28.28$\% \\
A100 & 128K & 42.037 s / 33.080 GiB & 43.791 s / 18.572 GiB & +4.17\% & $-43.86$\% \\
RTX 4090 & 32K & 5.500 s / 20.027 GiB & 6.191 s / 16.771 GiB & +12.55\% & $-16.26$\% \\
RTX 4090 & 64K & 15.698 s / 23.447 GiB & 15.865 s / 17.240 GiB & +1.06\% & $-26.47$\% \\
RTX 4090 & 128K & OOM & 47.110 s / 18.529 GiB & -- & feasible \\
Jetson & 32K & 116.659 s / 19.099 GiB & 126.492 s / 16.131 GiB & +8.43\% & $-15.54$\% \\
Jetson & 64K & OOM & 320.467 s / 16.631 GiB & -- & feasible \\
\bottomrule
\end{tabular}
\end{table*}

\textbf{Prefill cost and memory benefit.} Table~\ref{tab:latency-boundary} gives representative matched memory--latency points: at A100 128K, the process peak falls by 43.86\% for a 4.17\% TTFT increase; at RTX 4090 64K, it falls by 26.47\% for a 1.06\% increase.

\textbf{Decode behavior and timing composition.} The detailed measurements in Table~\ref{tab:overhead-details} show that the additional work is concentrated before generation. Across all six reported platform--length pairs, subsequent-token rates remain comparable or higher for TideKV-Snap; representative SnapKV/TideKV-Snap rates are 30.50/37.06 token/s on A100 at 64K, 27.13/31.59 token/s on RTX 4090 at 64K, and 2.64/3.45 token/s on Jetson at 32K. These paired measurements confirm that decoding from the matched compact cache remains intact. {\color{black} In separate profiling runs, layer-end scoring and compression account for 5.89\%--21.92\% of TideKV TTFT. These shares are medians of per-run ratios. They describe prefill composition, whereas Table~\ref{tab:latency-boundary} reports the TTFT increase over SnapKV.}

\begin{table}[pos=!t]
\centering
\scriptsize
\setlength{\tabcolsep}{3pt}
\caption{Decode rate and layer-end compression share. Decode entries are SnapKV/TideKV-Snap in token/s. {\color{black} Compression share is the median per-run ratio of compression time to TTFT from separate profiling runs (five on A100/RTX 4090; three on Jetson).}}
\label{tab:overhead-details}
\begin{tabular}{lrr}
\toprule
Platform, context & Decode rate & Compression share \\
\midrule
A100, 16K & 29.77 / 36.58 & {\color{black} 21.92\%} \\
A100, 64K & 30.50 / 37.06 & {\color{black} 13.38\%} \\
RTX 4090, 32K & 26.86 / 32.45 & {\color{black} 17.96\%} \\
RTX 4090, 64K & 27.13 / 31.59 & {\color{black} 13.73\%} \\
Jetson, 16K & 2.67 / 3.46 & {\color{black} 7.16\%} \\
Jetson, 32K & 2.64 / 3.45 & {\color{black} 5.89\%} \\
\bottomrule
\end{tabular}
\end{table}

\textbf{Chunk-size operating point.} Chunk size controls the memory--latency operating point. The cross-platform sweep in Table~\ref{tab:chunk-pareto} identifies $C=2048$ as a robust default: A100 gains modest latency at $C=4096$, whereas RTX 4090 and Jetson consume more memory without improving TTFT. {\color{black} A separate chunk-size sweep used Llama-3.1-8B-Instruct with a 4,096-token budget. As $C$ varied from 512 to 4,096, TideKV-Snap's 16-task LongBench mean ranged from 48.90 to 49.09.}

\begin{table}[pos=!t]
\centering
\scriptsize
\caption{TideKV-Snap chunk-size Pareto sweep. Each entry is TTFT in seconds / peak memory in GiB (NVML on A100/RTX 4090; CUDA-managed on Jetson). {\color{black} RTX 4090 and Jetson timings were measured independently of Table~\ref{tab:latency-boundary}.}}
\label{tab:chunk-pareto}
\resizebox{\columnwidth}{!}{%
\begin{tabular}{lccc}
\toprule
Platform and context & $C=1024$ & $C=2048$ & $C=4096$ \\
\midrule
A100, 64K & 16.783 / 17.125 & 14.464 / 17.283 & 13.899 / 17.713 \\
RTX 4090, 64K & 17.181 / 17.080 & 15.804 / 17.240 & 15.981 / 17.670 \\
Jetson, 32K & 128.27 / 16.037 & 125.66 / 16.131 & 128.62 / 16.320 \\
\bottomrule
\end{tabular}
}
\end{table}

Overall, TideKV provides substantial construction-memory headroom with acceptable prefill overhead across the tested platforms. Near device limits, this headroom enables much longer prompts while preserving the final cache size and stable decode-stage behavior.


\FloatBarrier
\Needspace{8\baselineskip}
\section{Conclusion}
\label{sec:conclusion}
TideKV constructs a compressed initial KV cache by streaming prompt chunks within each layer and delaying eviction until the layer has processed the complete prompt. This layer-major schedule preserves full-prompt-aware cache selection while limiting full-prompt KV residency to one active layer and transient activations to one chunk.

Matched experiments show that TideKV preserves the underlying selector's quality profile and final cache size while reducing the memory required to construct that cache. {\color{black} TideKV-Snap cuts the 64K transient construction footprint by 83.89\% on A100 and completes 128K inference on both an RTX 4090 and a Jetson AGX Orin; its RTX 4090 RULER score is 65.93.} The allocator, backend, and persistent workload results further show that the active-memory reduction holds across the tested runtime configurations. These results make layer-major delayed eviction a practical way to lower the prefill construction peak and extend feasible context length without reducing the final decode-cache budget. {\color{black} The remaining linear bridge and active-layer KV costs warrant further evaluation at larger batch sizes and beyond 128K.} Future work will integrate the layer-aware scheduler and layer-end cache transformation with continuous batching, prefix reuse, graph execution, and distributed serving, while exploring adaptive chunk sizing and backend-aware memory management. 











\printcredits

\section*{Acknowledgments}
This work is partially supported by the National Natural Science Foundation of China (No. 62472358, 62322601), and the Fundamental Research Funds for the Central Universities(No. G2026KY06261). 

\section*{Declaration of competing interest}
The authors declare that they have no known competing financial interests or personal relationships that could have appeared to influence the work reported in this paper.

\section*{Data availability}
Data will be made available on request.

\bibliographystyle{num}

\bibliography{cas-refs}



\end{document}